\documentclass[conference]{IEEEtran}
\IEEEoverridecommandlockouts

\usepackage[many]{tcolorbox}    	

\usepackage{pgfplots}
\pgfplotsset{compat=1.17}

\newtcolorbox{doubleLineBox}{
    enhanced, 
    boxrule = 0pt, 
    borderline = {0.75pt}{0pt}{main}, 
    borderline = {0.75pt}{2pt}{sub} 
}
\definecolor{main}{HTML}{FFFFFF}    
\definecolor{sub}{HTML}{cde4ff}     
\newtcolorbox{dottedBox}{
    colback = main,
    enhanced,
    boxrule = 1.5pt, 
    colframe = black, 
    left=0.2em,right=0.2em,top=0.2em,bottom=0.2em,
    borderline = {1.5pt}{0pt}{main, dashed} 
}

 \usepackage{pifont}

\usepackage[linesnumbered,ruled]{algorithm2e}
\usepackage{wrapfig}
\usepackage{hyperref}
\usepackage{subcaption} 
\usepackage{pifont}
\usepackage{enumitem}
\usepackage{tikz}
\usepackage{adjustbox}
\usepackage{multirow} 
\usepackage{colortbl}

\usepackage{cite}
\usepackage{amsmath,amssymb,amsfonts}
\usepackage{algorithmic}
\usepackage{graphicx}
\usepackage{textcomp}
\usepackage{xcolor}
\def\BibTeX{{\rm B\kern-.05em{\sc i\kern-.025em b}\kern-.08em
    T\kern-.1667em\lower.7ex\hbox{E}\kern-.125emX}}
\begin{document}

\title{Towards cost-effective and resource-aware aggregation at Edge for Federated Learning
\thanks{This work is sponsored in part by the NSF under the grants: CSR-2106634/2312785, CCF-1919113/1919075, OAC-2004751, and 3M.}
}

\author{\IEEEauthorblockN{1\textsuperscript{st} Ahmad Faraz Khan}
\IEEEauthorblockA{\textit{Virginia Tech} \\
Blacksburg, USA \\
ahmadfk@vt.edu}
\and
\IEEEauthorblockN{2\textsuperscript{nd} Yuze Li}
\IEEEauthorblockA{\textit{Virginia Tech} \\
Blacksburg, USA \\
lyuze@vt.edu}
\and
\IEEEauthorblockN{3\textsuperscript{rd} Xinran Wang}
\IEEEauthorblockA{\textit{University of Minnesota} \\
Minnesota, USA \\
wang8740@umn.edu}
\and
\IEEEauthorblockN{4\textsuperscript{th} 	Sabaat Haroon}
\IEEEauthorblockA{\textit{Virginia Tech} \\
Blacksburg, USA \\
sabaat@vt.edu}
\and
\IEEEauthorblockN{5\textsuperscript{th} 	Haider Ali}
\IEEEauthorblockA{\textit{Virginia Tech} \\
Blacksburg, USA \\
haiderali@vt.edu}
\and
\IEEEauthorblockN{6\textsuperscript{th} Yue Cheng}
\IEEEauthorblockA{\textit{University of Virginia} \\
Charlottesville, USA \\
mrz7dp@virginia.edu}
\and
\IEEEauthorblockN{7\textsuperscript{th} Ali R. Butt}
\IEEEauthorblockA{\textit{Virginia Tech} \\
Blacksburg, USA \\
butta@vt.edu}
\and
\IEEEauthorblockN{8\textsuperscript{th} Ali Anwar}
\IEEEauthorblockA{\textit{University of Minnesota} \\
Minnesota, USA \\
aanwar@umn.edu}
}

\newcommand{\yuze}[1]{\textcolor{cyan}{{[Yuze: #1]}}}
\newcommand{\ahmad}[1]{\textcolor{red}{{[Ahmad: #1]}}}
\newcommand{\anwar}[1]{\textcolor{green}{{[Anwar: #1]}}}
\newcommand{\faraz}[1]{\textcolor{blue}{{[Faraz: #1]}}}
\newcommand{\red}[1]{\textcolor{red}{{[#1]}}}
\newcommand{\chk}[1]{\textcolor{orange}{{#1}}}

\newcommand{\stepnum}[1]{\raisebox{.5pt}{\textcircled{\raisebox{-.9pt} {#1}}}}
\newcommand{\stepnumsimple}[1]{\textsf{#1)} \xspace}
\newcommand{\BULLET}{\vspace{+.00in} \noindent $\bullet$ \hspace{+.00in}}

\newcommand{\etc}{\textit{etc.}\xspace}
\newcommand{\ie}{\textit{i.e.,}\xspace}
\newcommand{\eg}{\textit{e.g.,}\xspace}
\newcommand{\etal}{\textit{et~al.}\xspace}
\newcommand{\wrt}{\textit{w.r.t.}\xspace}
\newcommand{\aka}{\textit{a.k.a.}\xspace}
\newcommand{\vs}{\textit{vs.}\xspace}

\newcommand{\mysection}[1]{
\section{#1}
}

\newcommand{\mysubsection}[1]{\vspace{-.00in}\subsection{#1}\vspace{-.00in}}
\newcommand{\mysubsubsection}[1]{\vspace{-.00in}\subsubsection{#1}\vspace{-.00in}}

\IEEEpubid{\makebox[\columnwidth]{979-8-3503-2445-7/23/\$31.00 ©2023 IEEE \hfill} 
\hspace{\columnsep}\makebox[\columnwidth]{ }}

\maketitle
\begin{abstract}
\begin{normalsize}

Federated Learning (FL) is a machine learning approach that addresses privacy and data transfer costs by computing data at the source. It's particularly popular for Edge and IoT applications where the aggregator server of FL is in resource-capped edge data centers for reducing communication costs. Existing cloud-based aggregator solutions are resource-inefficient and expensive at the Edge, leading to low scalability and high latency. To address these challenges, this study compares prior and new aggregation methodologies under the changing demands of IoT and Edge applications. This work is the first to propose an adaptive FL aggregator at the Edge, enabling users to manage the cost and efficiency trade-off. An extensive comparative analysis demonstrates that the design improves scalability by up to $4\times$, time efficiency by $8\times$, and reduces costs by more than $2\times$ compared to extant cloud-based static methodologies. \looseness=-1

\end{normalsize}
\end{abstract}

\mysection{Introduction}
\label{Introduction}
\vspace{3pt}

IoT and edge devices generate sensitive data that requires careful handling to prevent security or privacy violations. Traditional machine learning methods~\cite{abadi2016tensorflow,chilimbi2014project,ACADIA} that send private data to a centralized location for training pose significant privacy, communication, and security challenges and are unsuitable for sensitive data due to regulations like HIPAA~\cite{act1996health} and GDPR~\cite{regulation2018general}. Federated Learning (FL)~\cite{mcmahan2017communication} is a new technique that enables clients, such as IoT and edge devices, to collaborate in training a model without moving the data outside the system boundaries. FL reduces communication costs and privacy risks and has been adopted for a wide range of IoT and edge applications, including agriculture, healthcare, human behavior recognition, transportation, and smart homes~\cite{ MAROLI2021113488_agriculture, smart_healthcare, federated_transportation, smart_home}.

\textbf{Problem.} At the heart of Federated Learning (FL) lies the parameter/aggregator server, which coordinates the training process, manages client participation, and performs aggregation of model updates. However, when it comes to IoT and edge applications, this server must contend with unique scalability, efficiency, and cost management challenges. This is especially true given the scale of these applications, which often involve millions of client devices with highly segregated data~\cite{bonawitz2019towards, Oort}. As a result, addressing these challenges is critical to the success of FL in edge environments~\cite{advances_and_open_problems}. Edge data centers are compact data centers that have a smaller footprint compared to traditional data centers. They typically house only a limited number of servers, usually between 3 to 10, and can be easily moved and deployed in limited spaces such as a factory or an Internet Access point~\cite{gartner2022distributed}. For this reason, it becomes an attractive option to house aggregator servers at the edge to save communication costs~\cite{ federated_transportation}. The intuition of unlimited resource scalability in existing solutions for cloud aggregator servers becomes an impractical approach at the edge data center. Limited compute, network, and memory resources at the edge data centers raise the requirement for a resource-aware scalable, and efficient aggregator at the edge while maintaining minimal costs.

\textbf{Challenges.} Existing studies on FL aggregation services have primarily focused on different aspects such as communication efficiency~\cite{konecny2017federated}, time efficiency~\cite{bonawitz2019towards, fedn}, and cost reduction~\cite{fedless, lambdaFL}. However, these studies propose solutions for cloud settings where they typically assume no limitations on resources or costs for the aggregation server, therefore, these methodologies cannot be used for edge data center setups where resources are limited. Moreover, current FL frameworks~\cite{reina2021openfl, ludwig2020ibm, TnsorFlowFederated} rely on a single central node for the aggregator server, leading to inefficiencies, high I/O or communication costs, prolonged aggregation times, and limited scalability. Furthermore, there is a lack of understanding of the complex system challenges that arise when developing scientific FL applications, especially in the context of edge-cloud aggregation jobs~\cite{EdgeIoT,herring2020, hierarchical}. Thus, there is a need to develop new FL aggregation solutions that can effectively leverage edge data centers' resources while minimizing costs.

\textbf{Contributions.} In this paper, we aim to address the challenges of scalability, efficiency, and cost-effectiveness in FL aggregation at edge data centers. For this, we present a comprehensive analysis of the performance and cost of existing FL frameworks and multi-core, multi-node aggregation methodologies on limited resources, simulating conditions at the edge server. We acknowledge previous works such as serverless parameter servers~\cite{fedless, lambdaFL} and distributed aggregator servers~\cite{bonawitz2019towards}, but find that they have limited scalability and efficiency or assume unlimited cost spending. Additionally, some designs rely on peer-to-peer communication, which increases communication costs and latency. To address these challenges, we propose a resource-aware adaptive aggregator design that selects the most efficient methodology while keeping resource spending low.
Our design includes a Numba-based method, a Spark-based method~\cite{Zaharia10spark}, and a Serverless tree-reduce method, all integrated into a hybrid setting. This adaptive methodology enables the aggregator to scale according to demand while minimizing cost and latency. We use scalable storage for communication to overcome issues from peer-to-peer intra-aggregator connections. Additionally, users can customize the adaptive policy to achieve a balance between efficiency and cost trade-offs. In summary, this paper presents a novel resource-aware adaptive aggregator design for FL at edge data centers that address the challenges of scalability, time and resource efficiency, and cost-effectiveness. The goal is to reduce costs and spending on resources while maintaining the quality of service (QoS) indicated by low latency and cost-effective scalability for the user. To achieve this, the least costly method is chosen to meet the time efficiency and scalability requirements for particular aggregation jobs. Our approach differs from previous works by considering limited resources at the edge, and we believe this is the first work to do so.\looseness=-1

\textbf{Technical Insights.} We start by building a predictive model that outputs a methodology for aggregation, minimizing time and monetary costs based on user preferences, using exhaustive profiling data, task-specific information, and system availability as input. To assess the efficacy of the adaptive method, we perform a comprehensive analysis of each technique by highlighting its strengths and weaknesses and comparing it with the adaptive method. We use a client emulator with a large number of clients and perform extensive micro and macro benchmarks to analyze each technique. Our analysis reveals interesting trends, such as serverless being more efficient than Spark for processing data from a large number of clients but costly for other specific workloads compared to Spark. On the other hand, Spark remains cost-effective and efficient for processing larger data chunks. Thus, each method has its pros and cons. No single method can fully address all the challenges in aggregation at the Edge data center, leading to the development of the adaptive aggregator design. Existing works have been done in the cloud setting, where the aggregator has unlimited resources. However, these methodologies cannot be used for edge data center setups with limited resources, a significant challenge our method addresses. \looseness=-1


\textbf{Evaluation.} The evaluation consists of exhaustive experimentation with up to one million clients. Altogether, the duration of the experiments was approximately 1440 hours, and more than 100k lambda functions were used in total.

\textbf{Summary.} Our contributions are as follows:
\begin{enumerate}
    \item Highlight the \textbf{limitations of existing FL frameworks} by analysis of common basic operations in their aggregation.
    \item Propose and assess \textbf{three different strategies} for scalability, efficiency, and cost-effectiveness limitations.
    \item Propose the \textbf{first solution for an adaptive aggregator at the Edge}. Our adaptive aggregation technique is driven by user preferences to achieve high QoS based on popular FL edge and IoT applications.
    
    \item Provide a \textbf{client emulator} that connects to any cloud-based parameter server to conduct a cost-benefit analysis of static compared to the adaptive method.
    
    \item Offer \textbf{Counter-intuitive insights}, like the unexpectedly high costs of Serverless for specific workloads driving the need for resource-efficient and cost-effective aggregation.

    
    
    
\end{enumerate}

\mysection{Background \& Motivation}\label{sec:background}

In the FL process, devices train local models and share them with an Aggregator (central) server. This Aggregator combines these local models, creating a global model using algorithms like Federated Averaging~\cite{mcmahan2017communication}, Iterative Averaging, Gradient Aggregation, and Coordinate-wise median~\cite{pmlr-v80-yin18a_coordinate}. This global model goes back to clients for further training, repeating until the desired accuracy is reached. We use the IBM Federated Learning Library (IBMFL version 1.0.6)~\cite{ludwig2020ibm} as our baseline~\cite{ludwig2020ibm} referred to as Vanilla throughout the paper because it has a similar aggregator architecture as other common frameworks~\cite{abadi2016tensorflow, microsoftFLUTE}.
\textbf{We only consider synchronous aggregation as it has gained popularity in recent works and is more stable in terms of convergence~\cite{advances_and_open_problems, FL_survey, khan2023pifl, tiff}}.
Next, we highlight challenges faced by the aggregator server of IoT and Edge FL applications such as edge data center's limited resources including computing, communication, and energy, along with varying participation and device damage~\cite{IoT_breakdown,FL_defense_survey}. 



\textbf{Varying clients' participation:} In IoT and edge FL applications, client participation varies~\cite{federated_transportation, hetrogenousFL}, leading to varying model updates received by the aggregator. Conventional cloud servers cannot handle this in a resource-efficient manner as they are designed for unlimited scaling. An adaptive approach is needed to upscale for availability, and downscale for reducing cost, while maintaining efficiency.
\textbf{Varying model sizes:} Techniques like model pruning~\cite{adaptiveModelPruning}, sparsification~\cite{gradientCompressionQuantization}, and quantization~\cite{uplinkDownlinkQuantization} reduce communication costs for IoT. Models differ in size per client, posing a challenge for FL cloud solutions~\cite{ludwig2020ibm,abadi2016tensorflow, microsoftFLUTE} lacking resource-aware scaling. Advanced methods using salient parameters~\cite{SPATL} also cause varied model sizes per client device. Thus, an adaptive strategy is crucial to select the best aggregation method based on costs and resources.
\textbf{Multi-tenancy Absence:} Edge cloud's parameter and aggregator servers with multi-tenancy~\cite{2DFQ, workflow_scheduling} share limited resources at the edge data center. Existing FL aggregation techniques falter on Edge data centers because they assume unlimited resources for scaling. Thus, shifting to an efficient approach within confined resources is crucial.\looseness=-1


\textbf{Scaling for Dynamic Workloads:}
IoT and Edge FL applications depend on Edge data centers~\cite{EdgeIoT, MAROLI2021113488_agriculture, federated_transportation, smart_home, iot_survey} for aggregation. These applications have inconsistent participation rates due to limited resources~\cite{IoT_challenges}, and damage to client devices~\cite{IoT_breakdown}. Furthermore, Aggregator servers are kept close to IoT devices in Edge datacenters to cut communication cost. However, they face challenges such as limited resources while serving multiple applications. Existing FL aggregator designs~\cite{ludwig2020ibm, microsoftFLUTE, TnsorFlowFederated} lack the ability of resource-aware scaling for these variable workloads under limited resources. An adaptive aggregator is essential for flexible, resource-efficient scaling. 
\textbf{Improving Efficiency Performance:}
\label{Improving_efficiency_performance}
Aggregation is performed after each epoch in the training process. Therefore, efficient aggregation can benefit all FL applications by reducing the waiting times of idle client devices~\cite{smart_home, smart_healthcare, mobileNetworks} between training rounds, thus improving the quality of service (QoS) for clients and allowing them to utilize their resources for training. 
\textbf{User Autonomy for Controlling Costs:}
\label{ User_autonomy_for_controlling_costs}Edge centers manage many applications~\cite{EdgeIoT, Khalid2023}, and FL training is lengthy~\cite{advances_and_open_problems}. Saving energy and resources on edge aggregators is vital. A static approach might not scale well, costing more without downsizing.\looseness=-1

We propose an adaptive aggregator design that dynamically adjusts its scale based on incoming workload, preventing overspending on resources and catering to workload surges. This flexibility enables users to balance between efficiency and cost, adapting to changing device and workload dynamics while prioritizing cost savings.

\mysection{Methodologies and their Analysis}
\label{sec:Methodologies}

We present three diverse Federated Learning (FL) aggregation methods, leveraging technologies like Numba, Spark MapReduce, and Serverless with shared I/O channels, alongside implementation guidance and evaluations for limitations.\looseness=-1

\begin{table}{
\centering
\caption{Specifications of models}
\vspace{-8pt}
\begin{center}

\resizebox{0.8\columnwidth}{!}{%
\begin{tabular}{|c|c|p{0.3\columnwidth}|c|}
\hline
\textbf{\textit{Model}}& \textbf{\textit{Model Size}}& \textbf{\textit{Convolutional layers}} & \textbf{\textit{Dense layers}}\\
\hline
CNN4.6& 4.6 MB& 32, 64 & 128\\
\hline
CNN73& 73 MB& 32, 256, 512, 1024 & 128\\
\hline
CNN179& 179 MB& 32, 512, 1024, 1900 & 128\\
\hline
CNN239& 239 MB&32, 1024, 1900, 2400 & 128\\
\hline
CNN478& 478 MB& 32*2,1024*2, 1900*2, 2400*2 & 128*2\\
\hline
CNN717& 717 MB& 32*3, 1024*3, 1900*3, 2400*3 & 128*3\\
\hline
CNN956& 956 MB& 32*2,1024*2, 1900*2, 2400*2 & 128*4\\
\hline
Resnet50& 91 MB&~\cite{resnet50} &~\cite{resnet50}\\
\hline
VGG16& 528 MB&~\cite{vgg16} &~\cite{vgg16}\\
\hline
\end{tabular}
\label{table:benchmarks}
}
\end{center}
}
\vspace{-12pt}
\end{table}

\textbf{Focus of Analysis:} We analyze the conditions favoring each method's performance to inform the design of an adaptive aggregation policy balancing time and cost efficiency. Through experimental analysis, we seek to address these key questions:
\ding{172} \label{question:1} What are the precise gains achieved by parallel fusion algorithm computation? (Section ~\ref{Multi Core}) \ding{173} \label{question:2} How does horizontal scaling impact time efficiency and cost in multi-node methodologies, and what is the upper limit on number of clients? (Section ~\ref{Multi-Node Aggregation}) \ding{174} \label{question:3} How does the complexity of fusion algorithms affect memory and CPU bottlenecks in both multi-core and multi-node methods? (Sections ~\ref{Multi Core} \& ~\ref{Multi-Node Aggregation}) \ding{175} \label{question:4} What bottlenecks emerge in horizontally scalable multi-node fusion algorithms? (Section ~\ref{Multi-Node Aggregation}) \ding{176} Which method suits specific use cases outlined in Section ~\ref{sec:background}, and what are their performance and cost advantages? (Section ~\ref{sec:Methodologies})


\begin{figure*}[htbp]
\vspace{-1.0em}
\centering
\begin{subfigure}[htbp]{0.22\textwidth}
\centering
\includegraphics[width=1.00\textwidth]{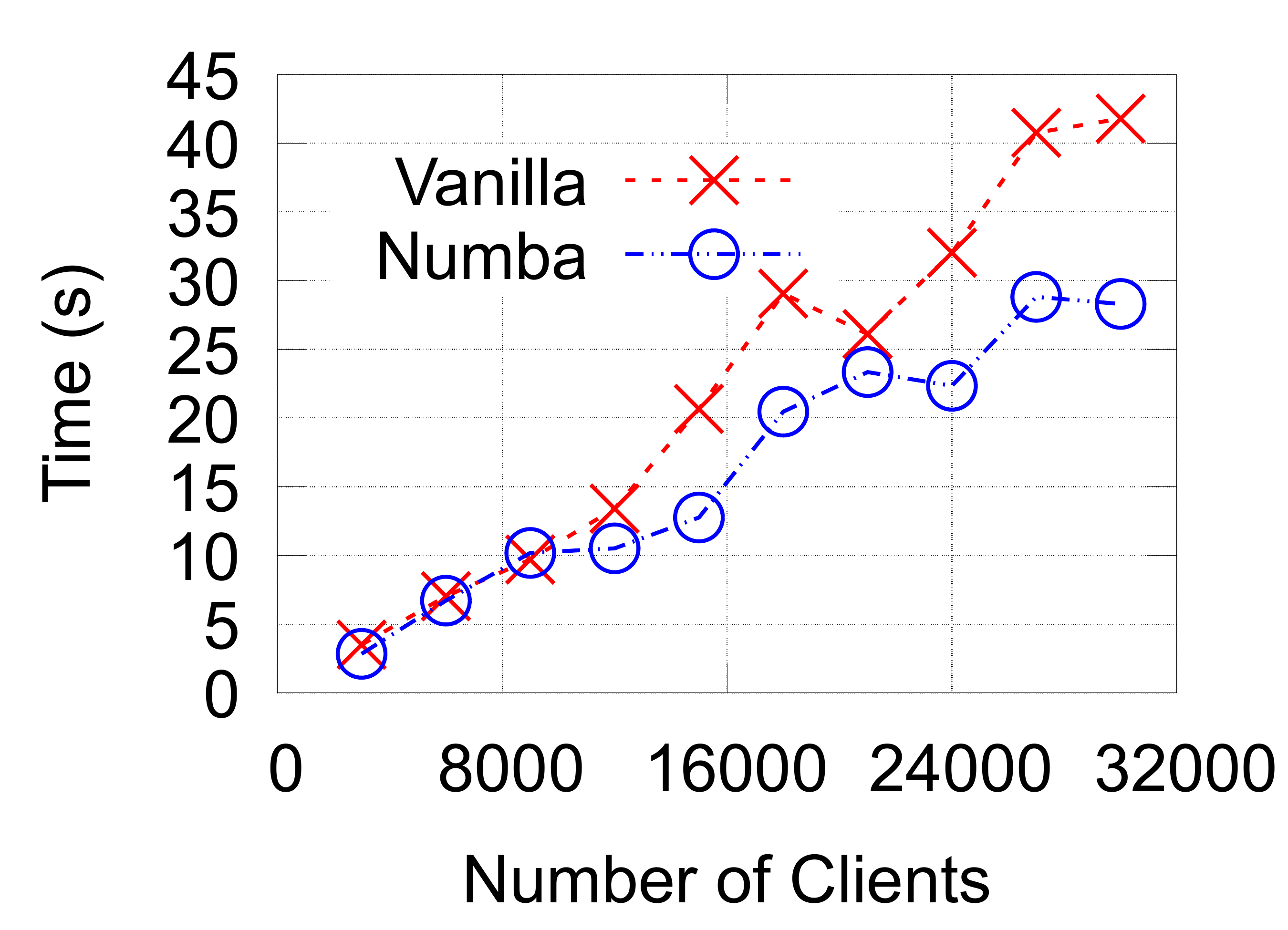}
\caption{Iteravg (CNN4.6)}
\label{fig:numpy_numba_comparison_4.6MB_iteravg}
\end{subfigure}
\begin{subfigure}[htbp]{0.22\textwidth}
\centering
\includegraphics[width=1.00\textwidth]{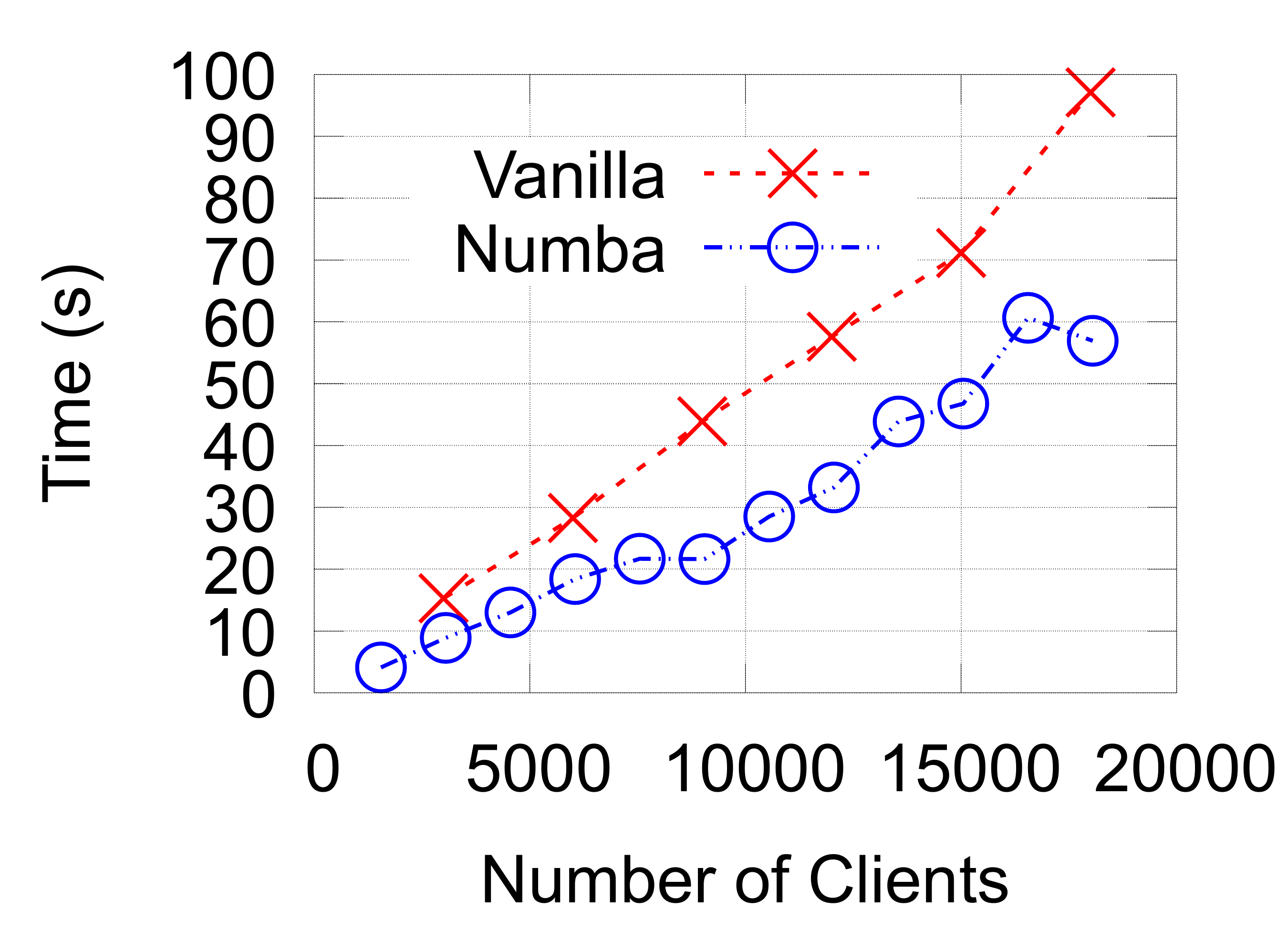}
\caption{Fedavg (CNN4.6)}
\label{fig:numpy_numba_comparison_4.6MB_fedavg}
\end{subfigure}
\begin{subfigure}[htbp]{0.22\textwidth}
\centering
\includegraphics[width=1.00\textwidth]{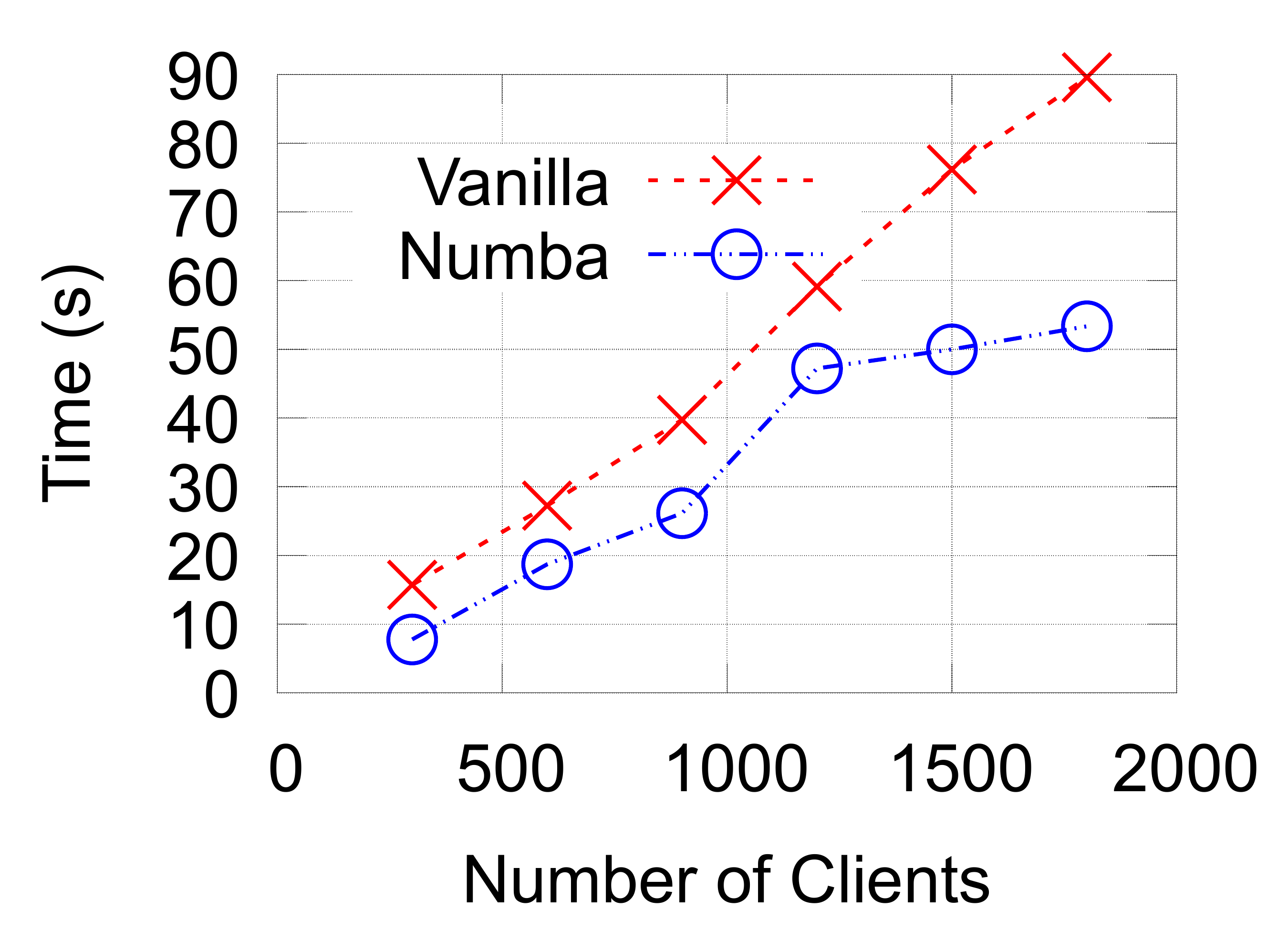}
\caption{Iteravg (Resnet50)}
\label{fig:numpy_numba_comparison_resnet50_iteravg}
\end{subfigure}
\begin{subfigure}[htbp]{0.22\textwidth}
\centering
\includegraphics[width=1.00\textwidth]{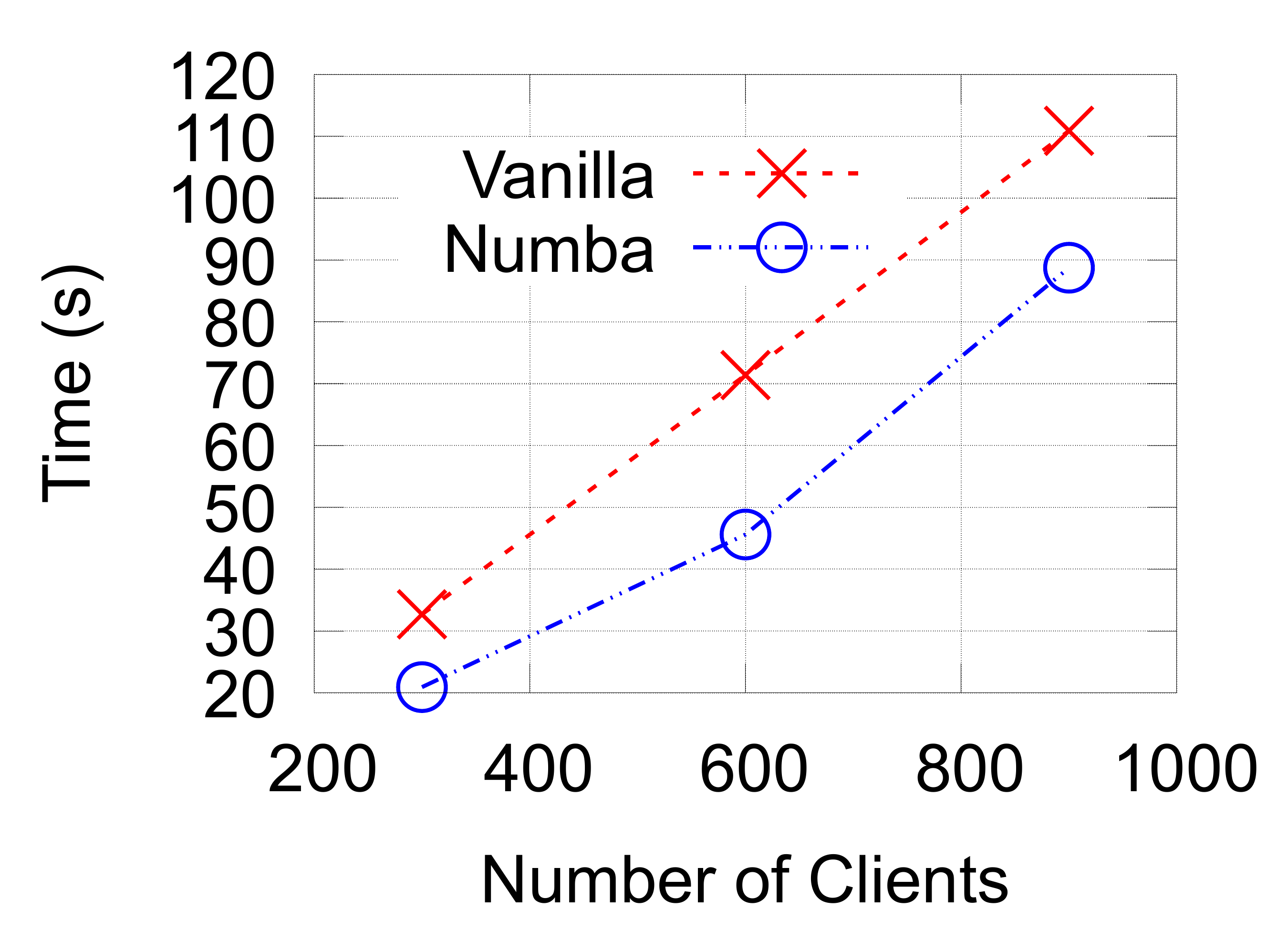}
\caption{Fedavg (Resnet50)}
\label{fig:numpy_numba_comparison_resnet50_fedavg}
\end{subfigure}
\vspace{-0.5em}
\caption{Average aggregation time comparison of Vanilla and Numba for smaller (CNN4.6) model and Resnet50 model}
\label{fig:numpy_numba_comparison_small_models}
\vspace{-.7em}
\end{figure*}

\mysubsection{Experimental Setup}
\label{sec:Experimental Setup}

\mysubsubsection{\textbf{Models}}
\label{Models}
We experimented with various CNN models, including Resnet50~\cite{resnet50} and VGG16~\cite{vgg16}, in different sizes listed in Table~\ref{table:benchmarks}, systematically increasing model sizes for comprehensive analysis, with model sizes comparable to \textbf{MobileNetV2 (4.2 MB)}, \textbf{ShuffleNetV2 (70 MB)}, \textbf{ResNet50 (224 MB)}, \textbf{InceptionV3 (450 MB)}, and \textbf{VGG19 (518 MB)}.

\mysubsubsection{\textbf{Client Emulator}}
\label{ClientEmulator}
The client emulator streamlines client simulation for researchers by simplifying technical intricacies. It comprises three stages, utilizing AWS S3: first, it sends an HTTP request to AWS S3 to move files from client to aggregator buckets, adjusting to regions and model quantities. Then, it uses monitoring to wait for the aggregated model in the aggregator's bucket. Finally, it triggers multiple S3 requests to return the model to the clients' bucket. Furthermore, the client emulator allows the simulation of user-defined randomized dropouts.\looseness=-1

\mysubsubsection{\textbf{Testbed}}
\label{subsec:Testbed}
\textbf{Spark-based Methodology:}
We evaluated the Spark-based method using a Spark cluster with 256 cores and 452 GB of memory. Apache Spark~\cite{Zaharia10spark} version 3.2.0 ran on Apache Hadoop Yarn~\cite{Vavilapalli2013ApacheHY_yarn} version 3.2.2. Executors in Spark had a 35 GB memory limit. We used HDFS with 2.6 TB storage for model updates, and executor container specifications were adjusted based on workload.
\textbf{Serverless Methodology:}
For the Serverless method, we employed AWS Lambda with a 4 GB memory limit per function.
\textbf{Numba-based Methodology:}
To compare the Numba-based method, we used a containerized setup on a single node with 64 cores, 256 GB memory, and 10 Gbit/s network bandwidth. The client emulator from Section~\ref{ClientEmulator} was used, with a five percent client sample in all experiments. Our analysis design was implemented in around 2K lines of Python code.\looseness=-1

\mysubsubsection{\textbf{Metrics}}
\label{Metrics}
We assessed four key metrics to evaluate the efficiency of each methodology: \stepnum{1} Client sampling rate for aggregation per round (default at 5\% of total clients), indicating scale achieved. \stepnum{2} Time for aggregation, assessing its efficiency. \stepnum{3} Aggregation cost, reflecting the value for money spent. \stepnum{4} Communication time, representing latency in model updates between clients. We also measured client write and read efficiency from the scalable storage used by the aggregator in our client emulator. These metrics collectively offer insights into the resource efficiency of each approach.

\mysubsubsection{\textbf{Fusion Algorithms}}
\label{Fusion Algorithms}

We analyzed two fundamental averaging-based fusion algorithms: Federated averaging (FedAvg) and Iterative averaging (IterAvg) from Vanilla. These serve as the basis for various fusion methods, including ClippedAveraging, ConditionalThresholdAveraging (discussed in~\cite{reina2021openfl}), and Gradient Aggregation algorithms (mentioned in~\cite{ludwig2020ibm}). FedAvg, defined in Equation~\ref{eq:fedavg}, employs client weights ($w_{i}$) with a total client count ($n_{total}$) and a small $\epsilon$ value of $10^{-6}$. Additionally, we are exploring more advanced fusion techniques like Zeno~\cite{zeno} and Coordinate-wise median~\cite{pmlr-v80-yin18a_coordinate}.

\vspace{-8pt}
\begin{equation}
\label{eq:fedavg}
    M = \sum_{i=1}^{n}  w_{i}/ (n_{total} + \epsilon)
\end{equation}
\vspace{-12pt}

\mysubsection{Multi-core Aggregation}
\label{Multi Core}

Our FL aggregation was optimized by switching from Numpy~\cite{harris2020arraynumpy} to Numba~\cite{numba}, enhancing parallel processing. This change, implemented in IBMFL’s FedAvg and IterAvg classes, retained the original communication and client training functions. It enabled storing client updates in the aggregator's memory for quicker, more efficient aggregation, especially beneficial for IoT applications with simpler models. This transition to Numba required only minor code modifications within the FL framework.


To answer question~\stepnum{1} in section~\ref{sec:background}, we created micro-benchmarks to analyze the Numba-based aggregation method compared to the Vanilla implementations of FedAvg and IterAvg algorithms. Figures~\ref{fig:numpy_numba_comparison_resnet50_iteravg} and~\ref{fig:numpy_numba_comparison_resnet50_fedavg} show the aggregation time comparison for the Resnet50 model. For Fedavg, we observed a $40.44\%$ reduction in execution time using the Numba-based method with 1.8k clients. However, due to fixed single-node memory, the number of clients supported by Resnet50 is lower compared to a smaller model like CNN4.6, which is why the Numba-based method performs similarly to Vanilla for the Resnet50 model. With higher participation, the Numba-based method can parallelize computations and improve time efficiency. This is evident in Figures~\ref{fig:numpy_numba_comparison_4.6MB_iteravg} and~\ref{fig:numpy_numba_comparison_4.6MB_fedavg} for the CNN4.6 model, where the Numba-based method reduced aggregation time by $56.85\%$. Moreover, the efficiency of the Numba-based method depends on the fusion algorithm used. Numba parallelizes loops to compute weighted averages in Fedavg, resulting in greater efficiency, while simpler calculation in Iteravg means fewer efficiency gains from parallel computation. \looseness=-1

In summary, the Numba-based method outperforms IBMFL and other FL frameworks that use the Numpy library for implementing fusion algorithms when client participation rates increase. It is also more resource-efficient as it utilizes all available cores to maintain efficiency reducing resource idling. However, for a smaller number of clients, parallel processing is less beneficial. These insights can be extended to other frameworks such as TensorFlow~\cite{abadi2016tensorflow}, Microsoft FLUTE~\cite{microsoftFLUTE}, and FLOWER~\cite{beutel2022flower}. This analysis also suggests that multiple resources (CPU, Memory) act as bottlenecks during aggregation. We also note that when we move from IterAvg to FedAvg, which is a slightly more complex algorithm, the CPU becomes a bottleneck. We conclude that the \textbf{CPU bottleneck is directly related to the complexity of the algorithm}, and as we move to more complex algorithms, the bottleneck induced by CPU resources will increase, answering question~\stepnum{3} raised in section~\ref{sec:background}. Considering this result, we argue that \textbf{specialized hardware (GPU)} as a solution cannot fully resolve multiple bottlenecks and can be expensive. This motivates us to explore and analyze further aggregation methodologies that can be used for larger workloads using CPU.\looseness=-1

\begin{figure}
\centering
\vspace{-1.0em}
\centerline{\includegraphics[width=.89\columnwidth]{ 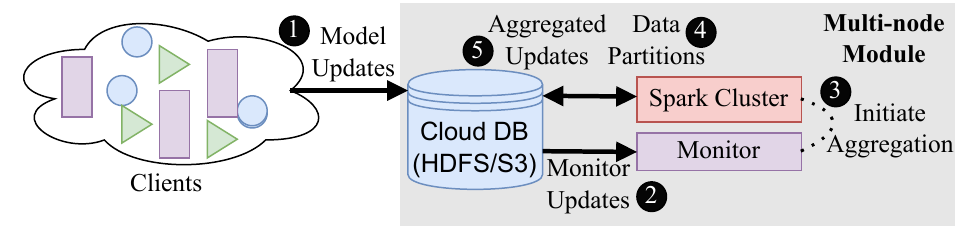}}
\vspace{-7pt}
\caption{Multi-node module design}
\label{fig:distributed_architecture}
\vspace{-0.5em}
\end{figure}

\begin{figure}[htbp]
\vspace{-1.5em}
\centering
\begin{subfigure}[htbp]{0.49\columnwidth}
\centering
\includegraphics[width=1.00\columnwidth]{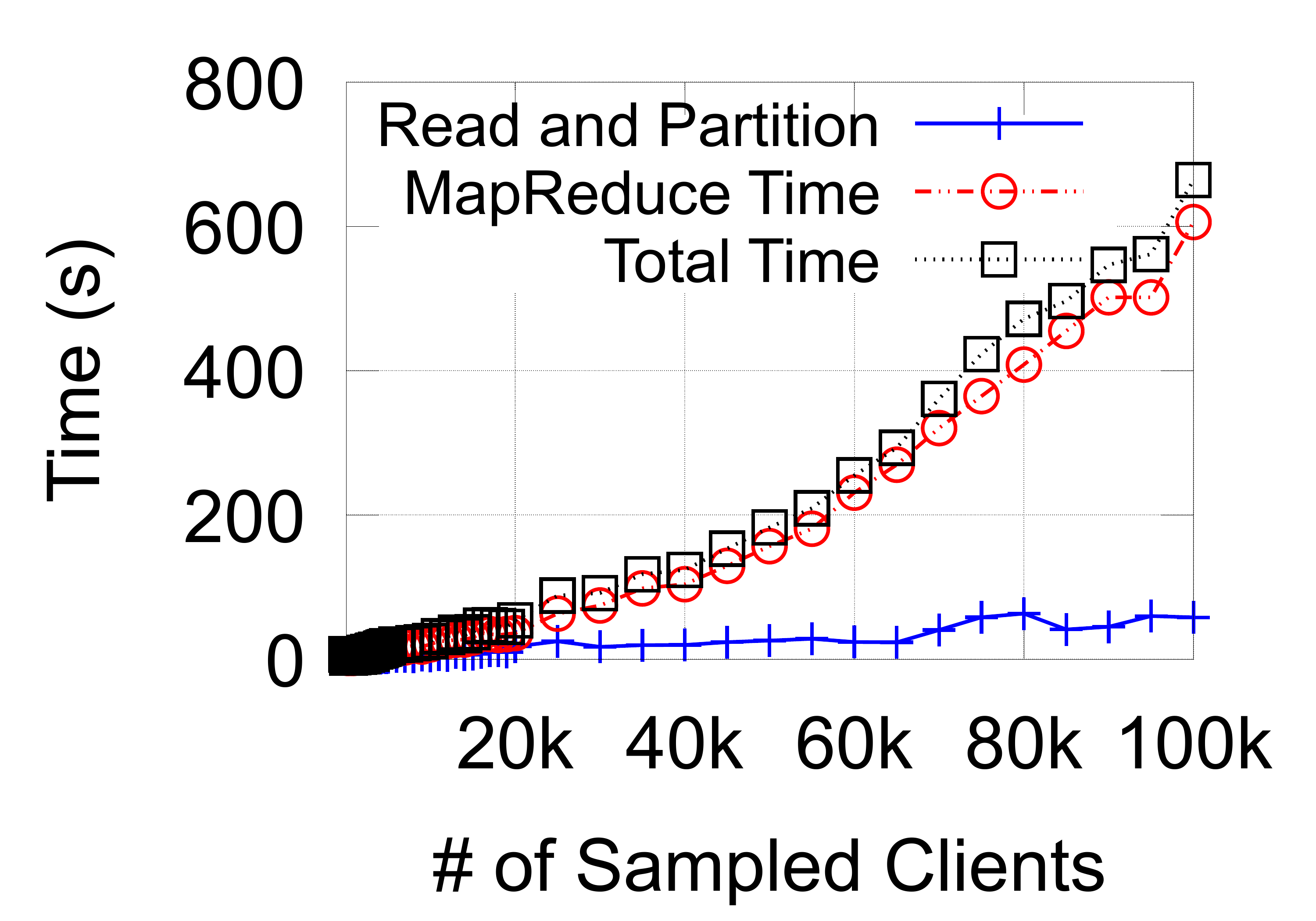}
\vspace{-15pt}
\caption{Fedavg}
\label{fig:pyspark_fedavg_4.6MB}
\end{subfigure}
\begin{subfigure}[htbp]{0.49\columnwidth}
\centering
\includegraphics[width=1.00\columnwidth]{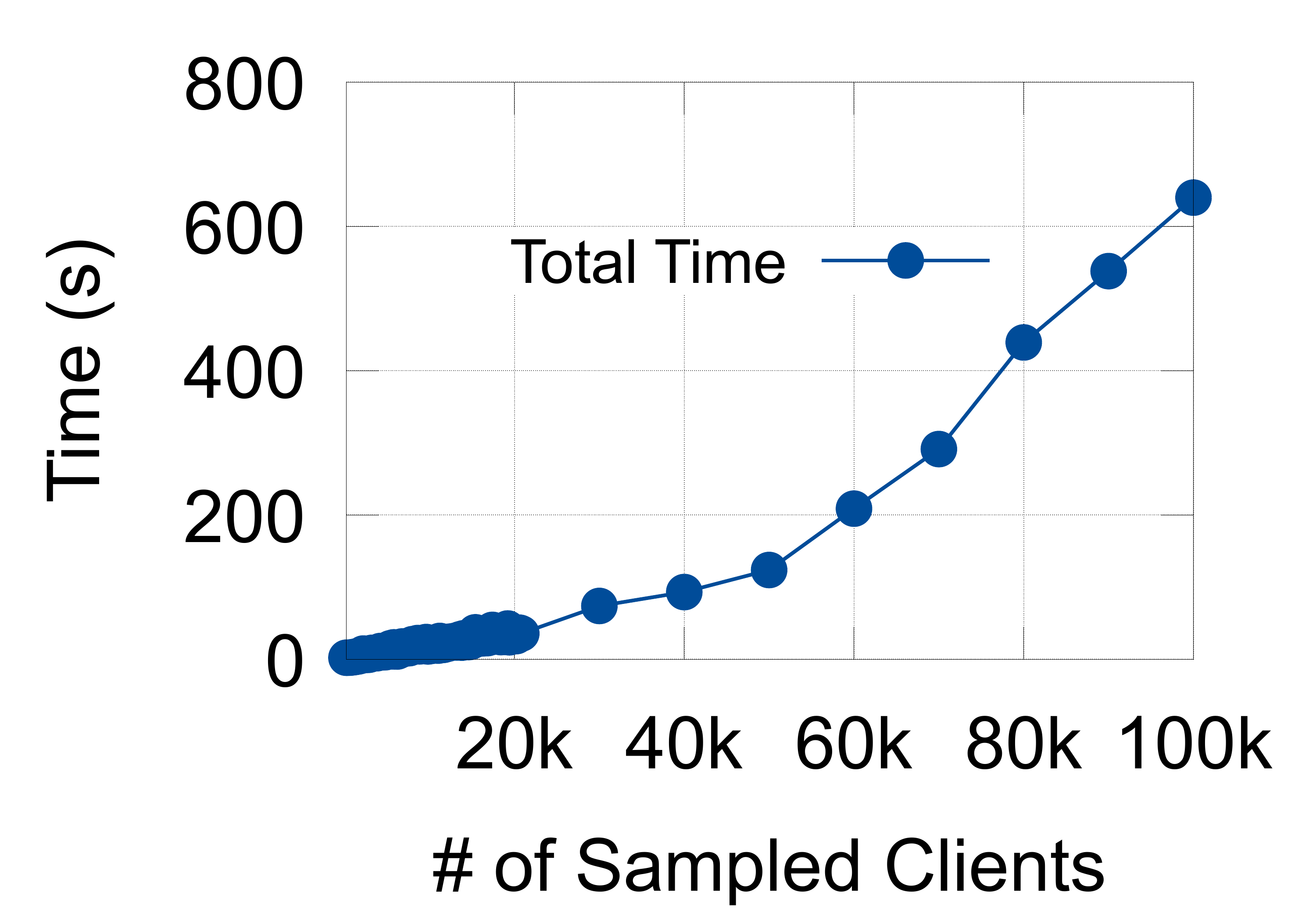}
\vspace{-15pt}
\caption{Iteravg}
\label{fig:pyspark_iteravg_4.6MB}
\end{subfigure}
\vspace{-0.8em}
\caption{Average aggregation time comparison of the Spark method with Fedavg and Iteravg on the CNN4.6 model}
\label{fig:pyspark_federated_and_iteravg_4.6}
\vspace{-10pt}
\end{figure}

\begin{figure*}[htbp]

\centering
\begin{subfigure}[htbp]{0.21\textwidth}
\centering
\includegraphics[width=1.00\textwidth]{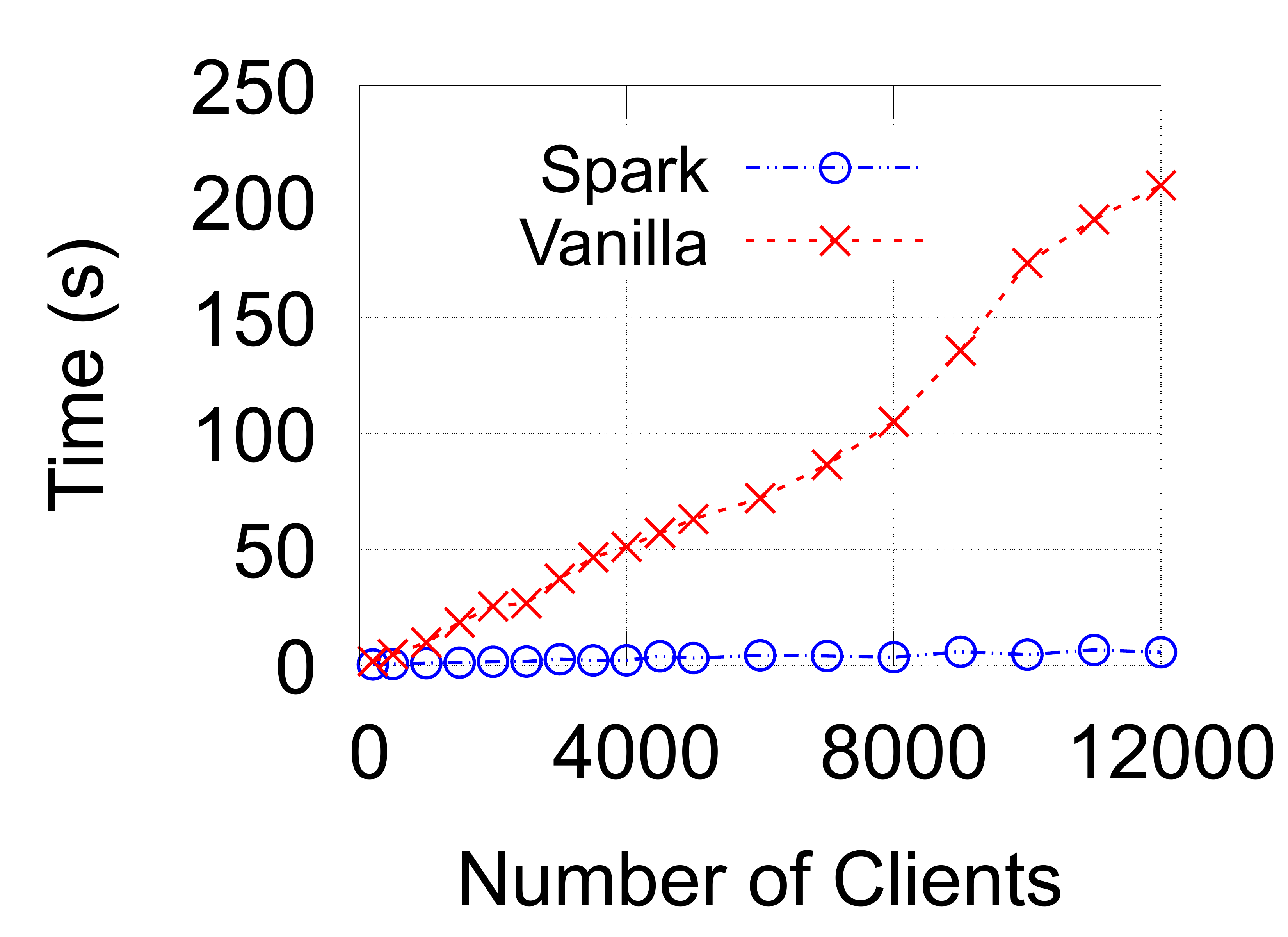}
\caption{CNN4.6}
\label{fig:pyspark_fedavg_4.6MB_pyspark_numpy_comparison_compressed}
\end{subfigure}
\begin{subfigure}[htbp]{0.21\textwidth}
\centering
\includegraphics[width=1.00\textwidth]{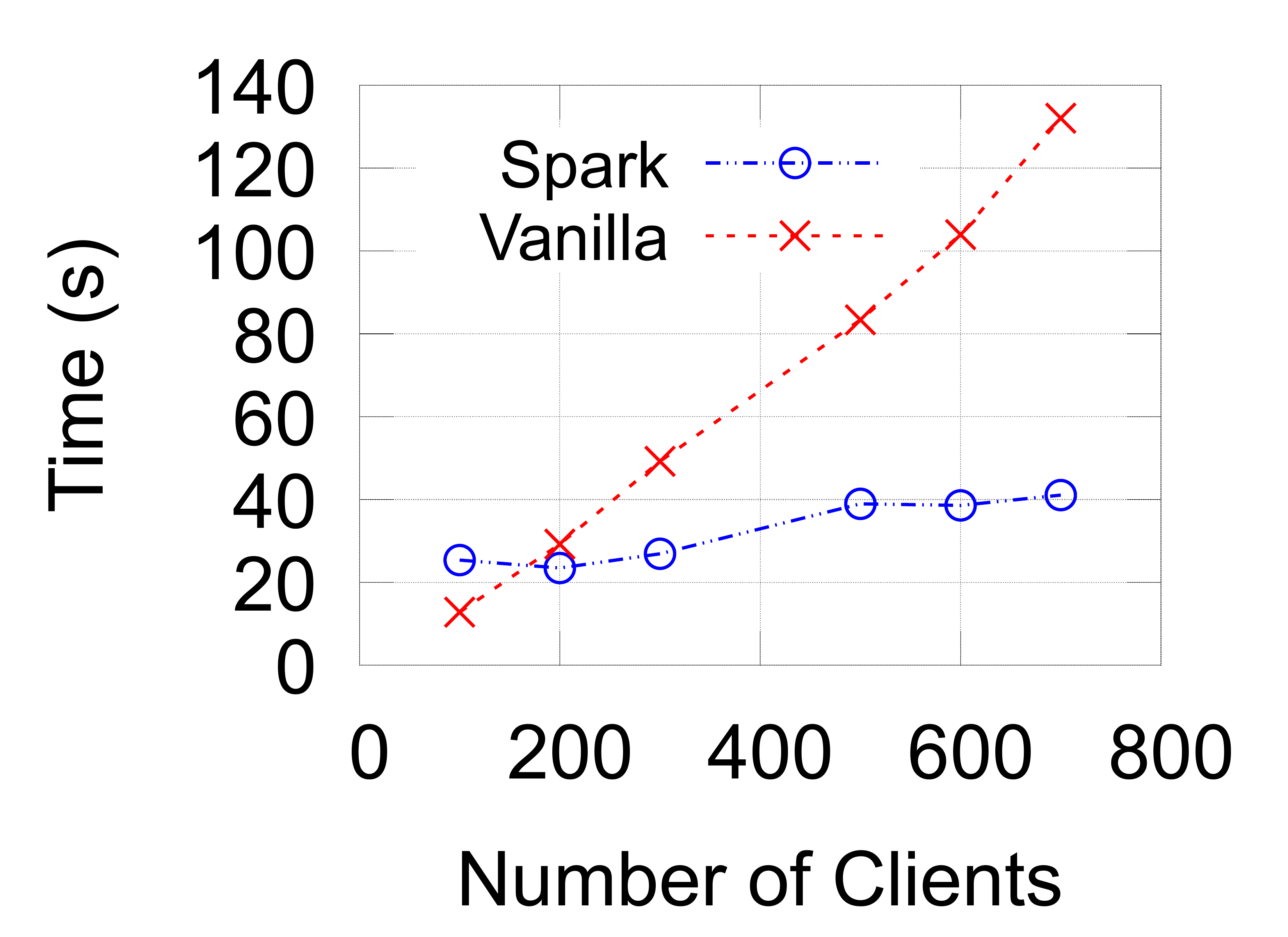}
\caption{CNN73}
\label{fig:pyspark_fedavg_73MB_pyspark_numpy_comparison_compressed}
\end{subfigure}
\begin{subfigure}[htbp]{0.21\textwidth}
\centering
\includegraphics[width=1.00\textwidth]{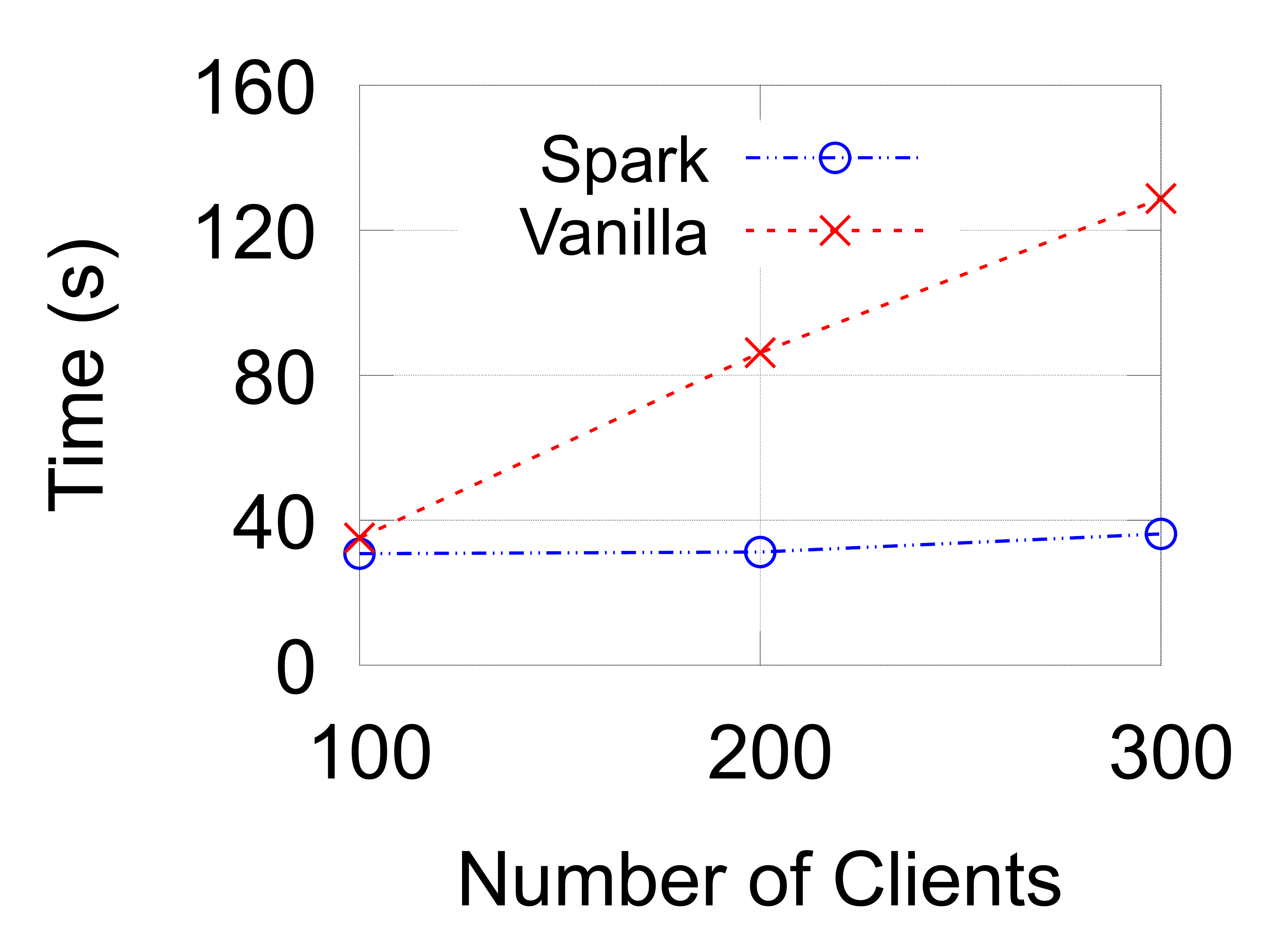}
\caption{CNN179}
\label{fig:pyspark_fedavg_179MB_pyspark_numpy_comparison_compressed}
\end{subfigure}
\begin{subfigure}[htbp]{0.21\textwidth}
\centering
\includegraphics[width=1.00\textwidth]{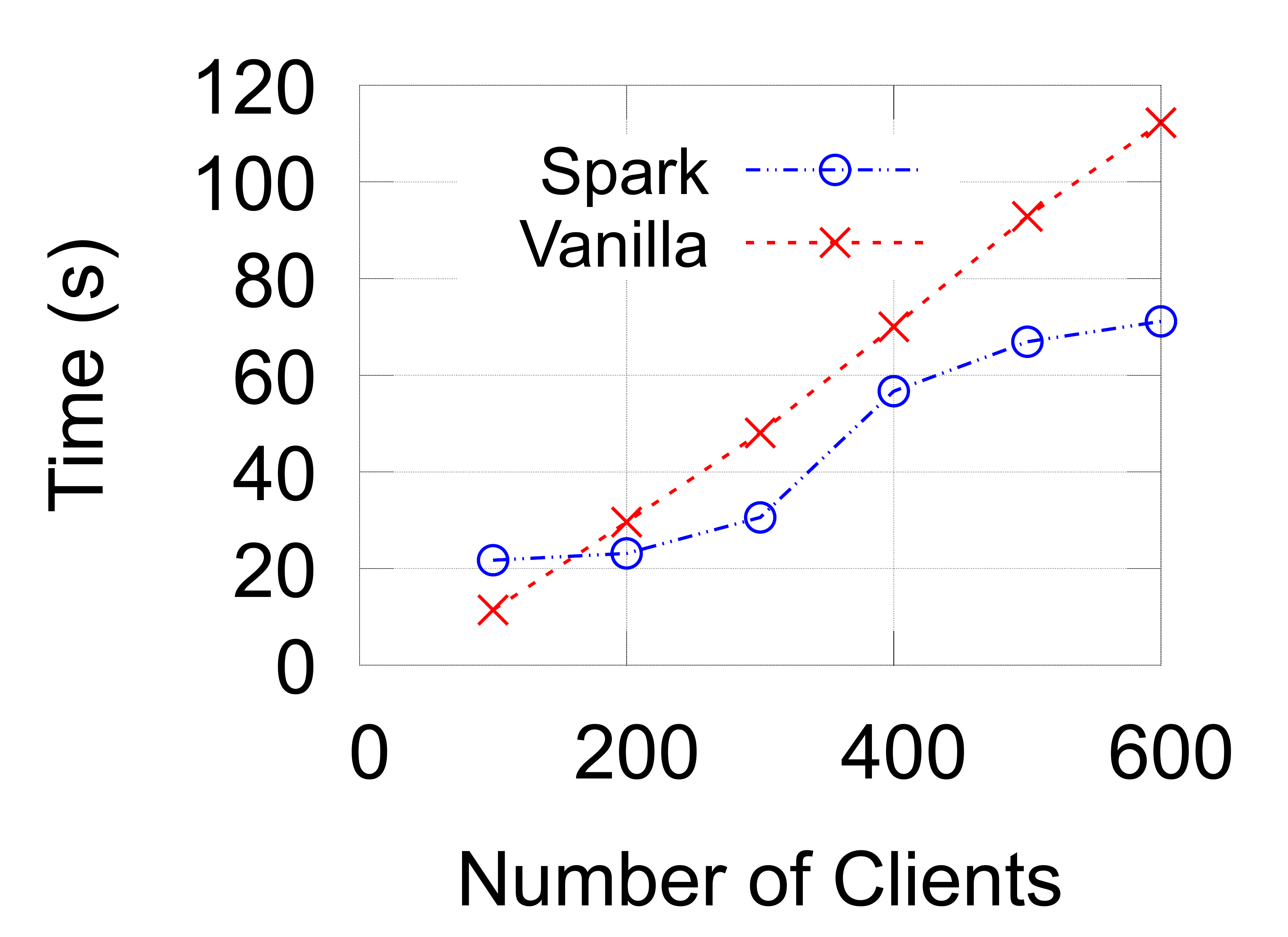}
\caption{Resnet50}
\label{fig:pyspark_fedavg_resnet50_pyspark_numpy_comparison_compressed}
\end{subfigure}
\vspace{-0.7em}
\caption{Average aggregation time comparison for Vanilla and Spark-based method for FedAvg with model compression}
\label{fig:pyspark_numpy_comparison_fedavg}
\vspace{-1.0em}
\end{figure*}

\begin{figure*}
    
\centering
\begin{subfigure}[htbp]{0.35\columnwidth}
\centering
\includegraphics[width=1.00\columnwidth]{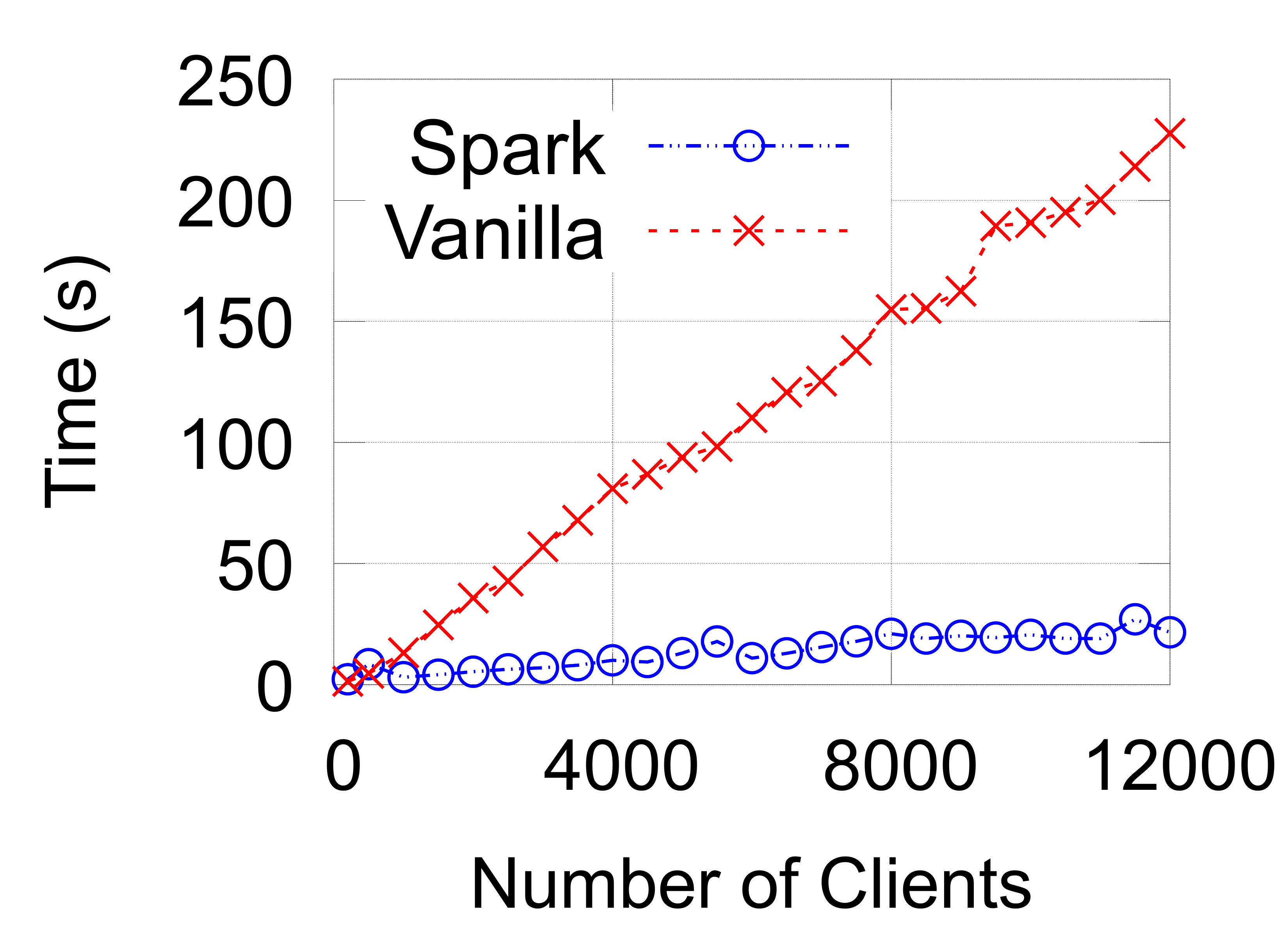}
\caption{CNN4.6}
\label{fig:pyspark_iteravg_4.6MB_pyspark_numpy_comparison_compressed}
\end{subfigure}
\begin{subfigure}[htbp]{0.35\columnwidth}
\centering
\includegraphics[width=1.00\columnwidth]{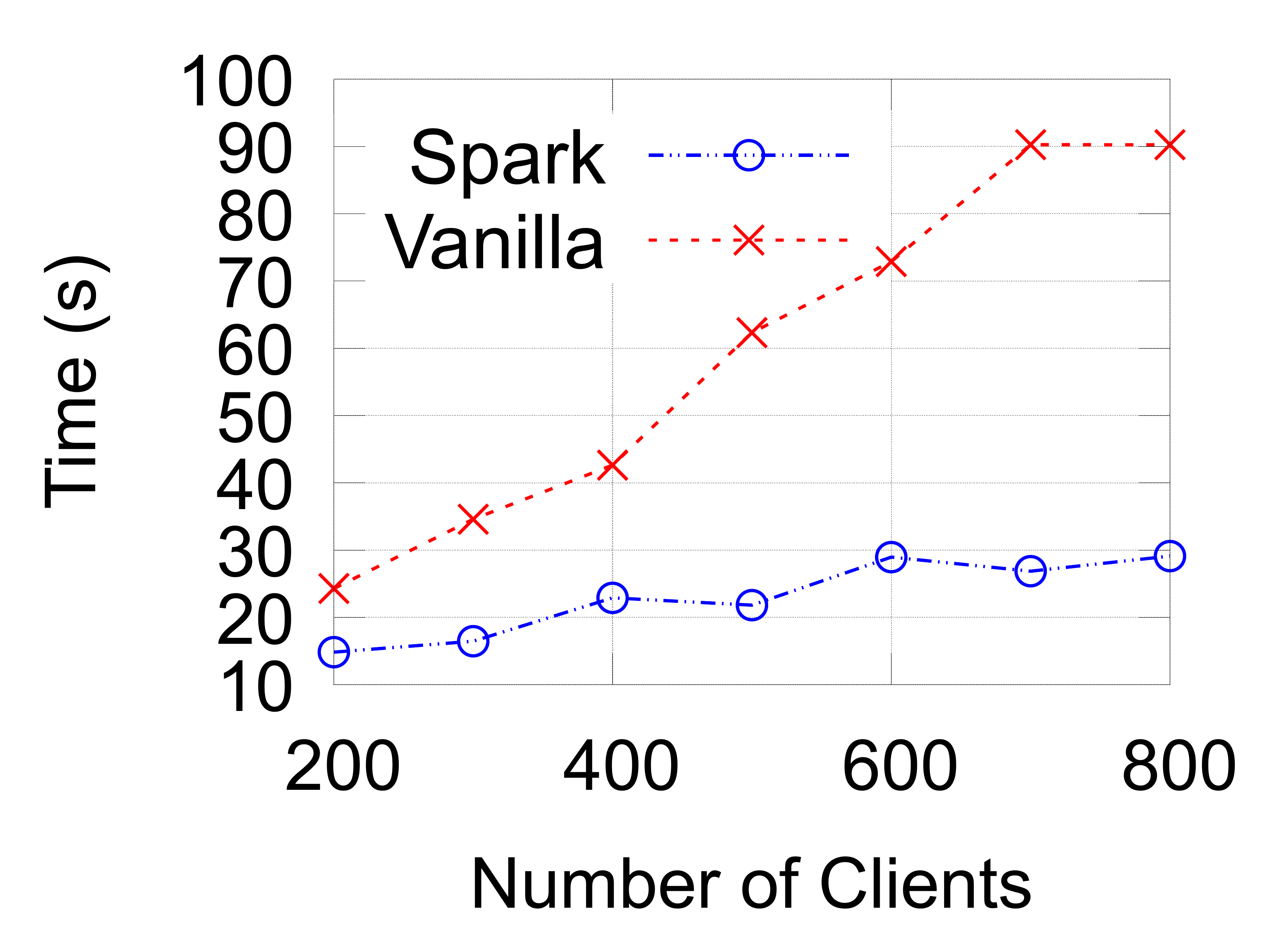}
\caption{CNN73}
\label{fig:pyspark_iteravg_73MB_pyspark_numpy_comparison_compressed}
\end{subfigure}
\begin{subfigure}[htbp]{0.35\columnwidth}
\centering
\includegraphics[width=1.00\columnwidth]{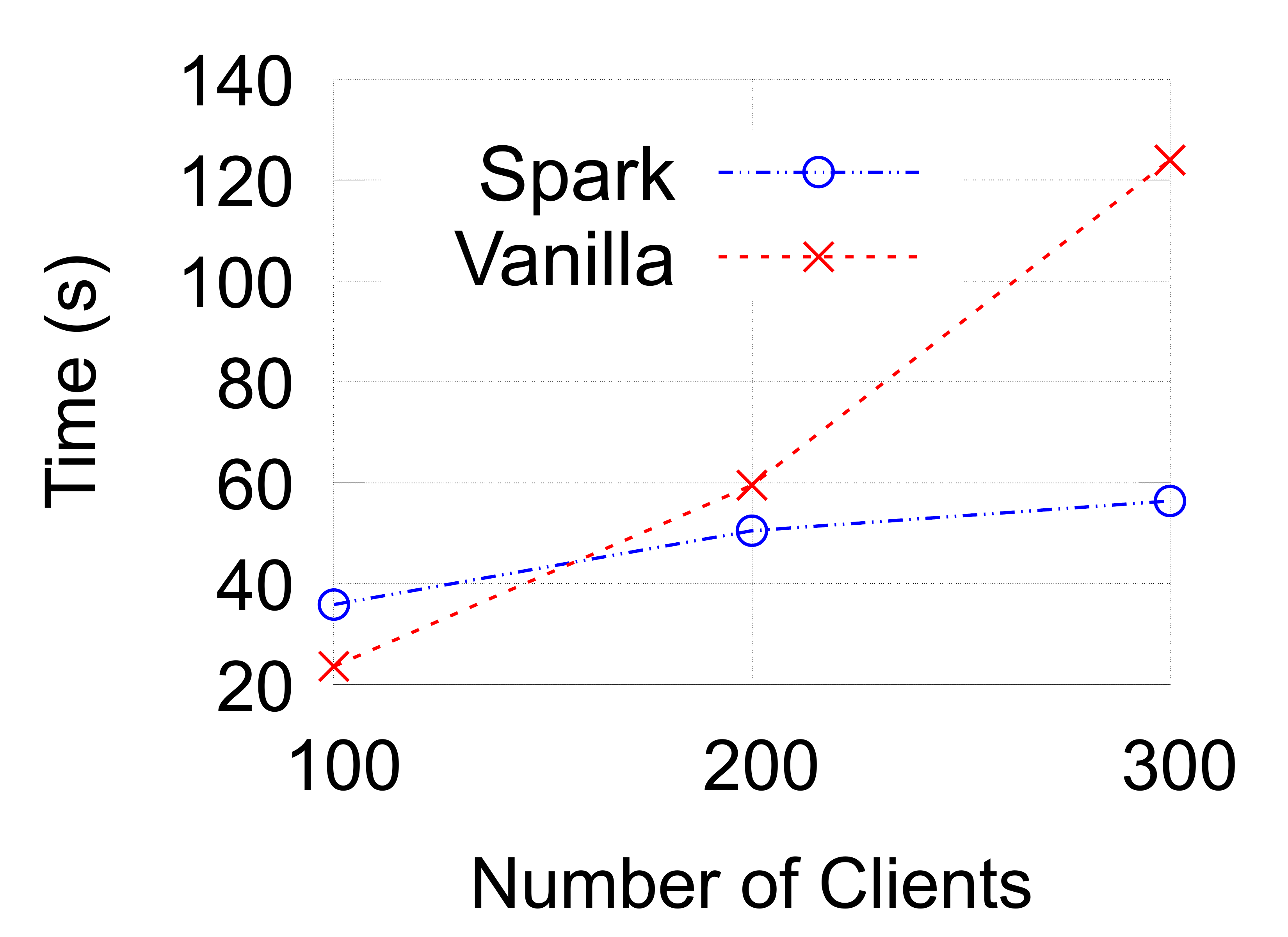}
\caption{CNN179}
\label{fig:pyspark_iteravg_179MB_pyspark_numpy_comparison_compressed}
\end{subfigure}
\begin{subfigure}[htbp]{0.35\columnwidth}
\centering
\includegraphics[width=1.00\columnwidth]{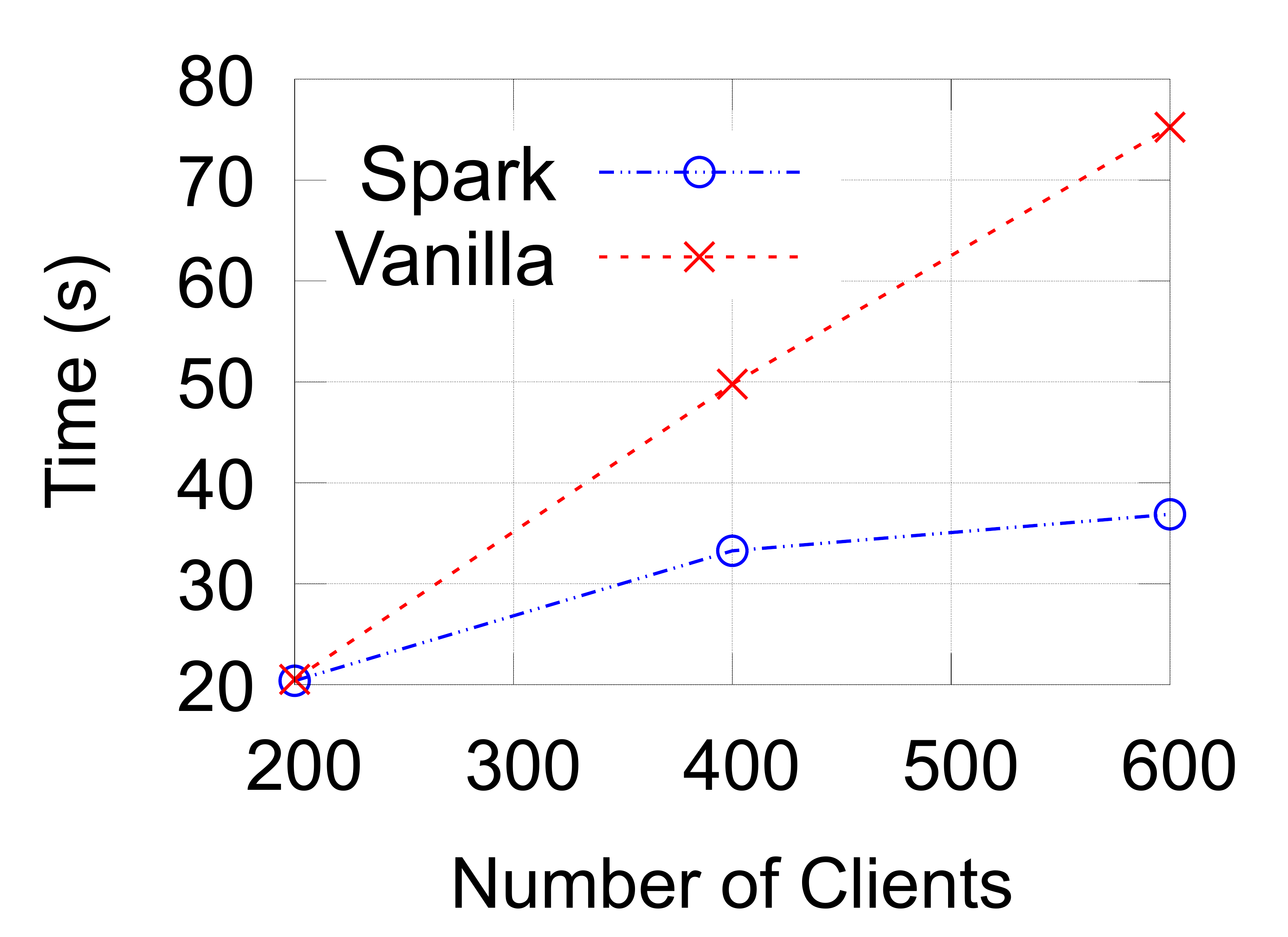}
\caption{Resnet50}
\label{fig:pyspark_iteravg_resnet50_pyspark_numpy_comparison_compressed}
\end{subfigure}
\begin{subfigure}[htbp]{0.35\columnwidth}
\centering
\includegraphics[width=1.00\columnwidth]{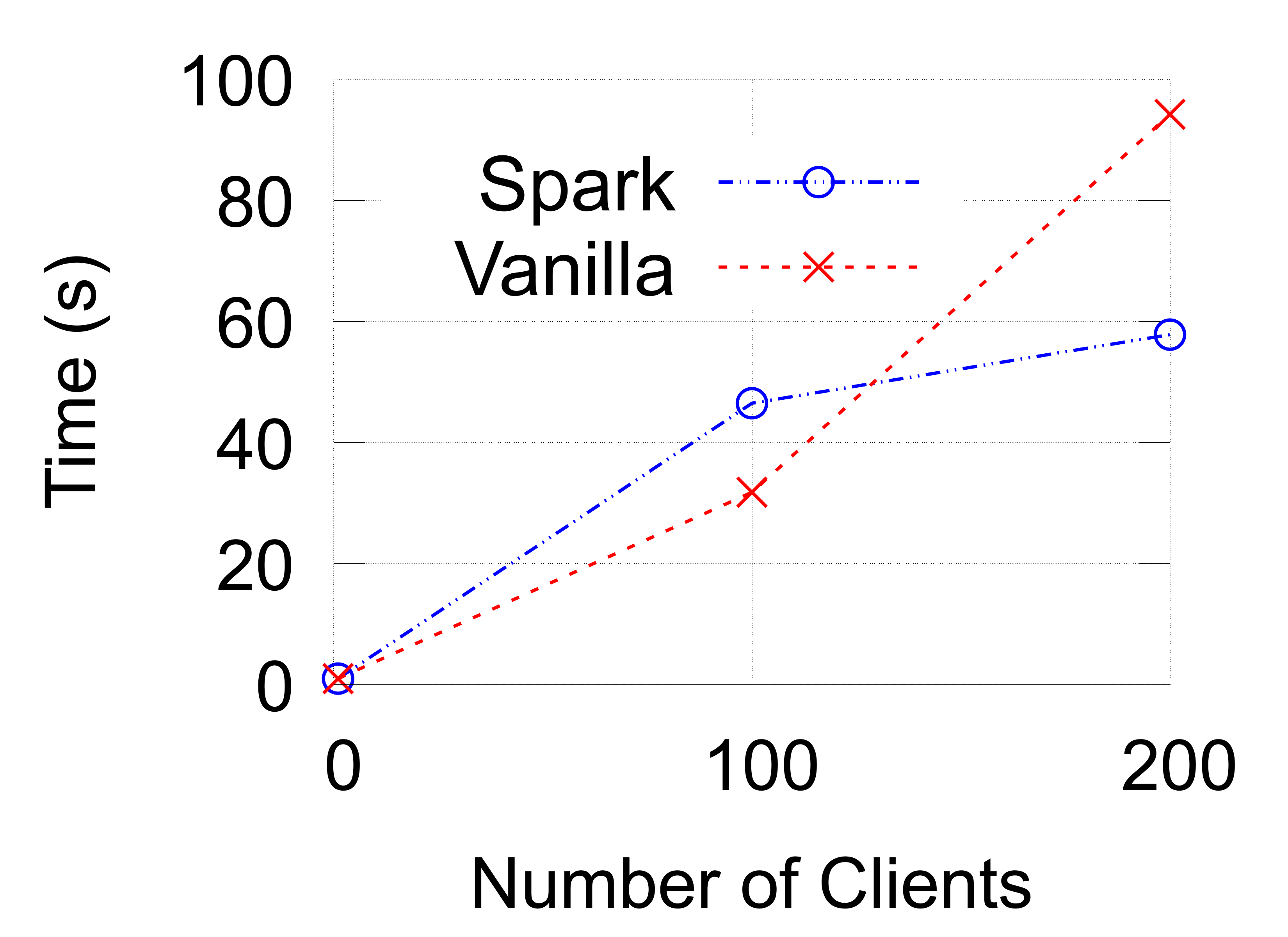}
\caption{CNN239}
\label{fig:pyspark_iteravg_239MB_pyspark_numpy_comparison_compressed}
\end{subfigure}
\vspace{-0.7em}
\caption{Average aggregation time comparison for Vanilla and Spark-based method using for IterAvg with model compression}
\vspace{-1.0em}
\label{fig:pyspark_numpy_comparison_iteravg_compressed}
\end{figure*}

\begin{figure*}[htbp]
\centering

\begin{subfigure}[htbp]{0.21\textwidth}
\centering
\includegraphics[width=1.00\textwidth]{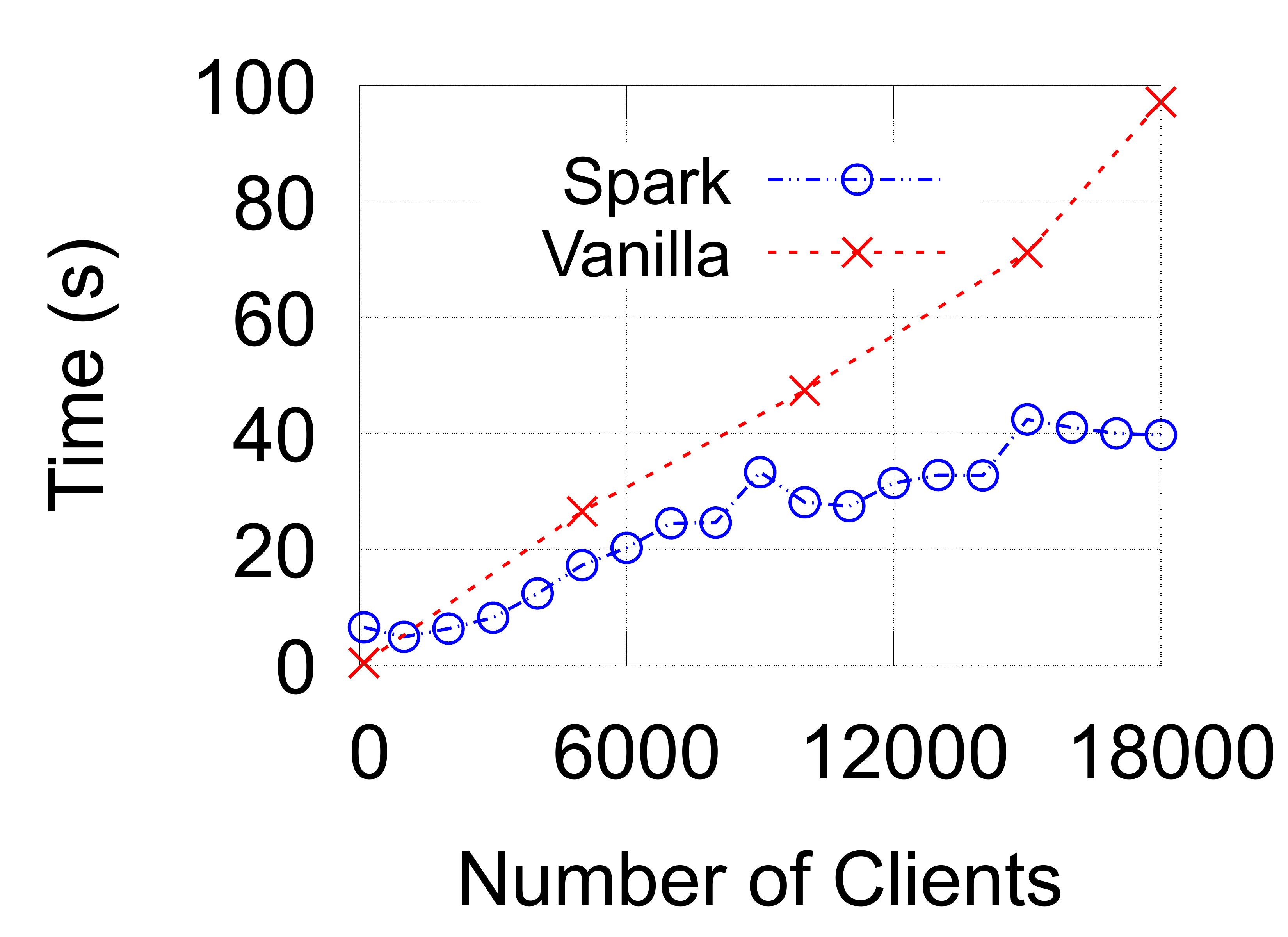}
\caption{CNN4.6}
\label{fig:pyspark_fedavg_4.6MB_pyspark_numpy_comparison_no_compressed}
\end{subfigure}
\begin{subfigure}[htbp]{0.21\textwidth}
\centering
\includegraphics[width=1.00\textwidth]{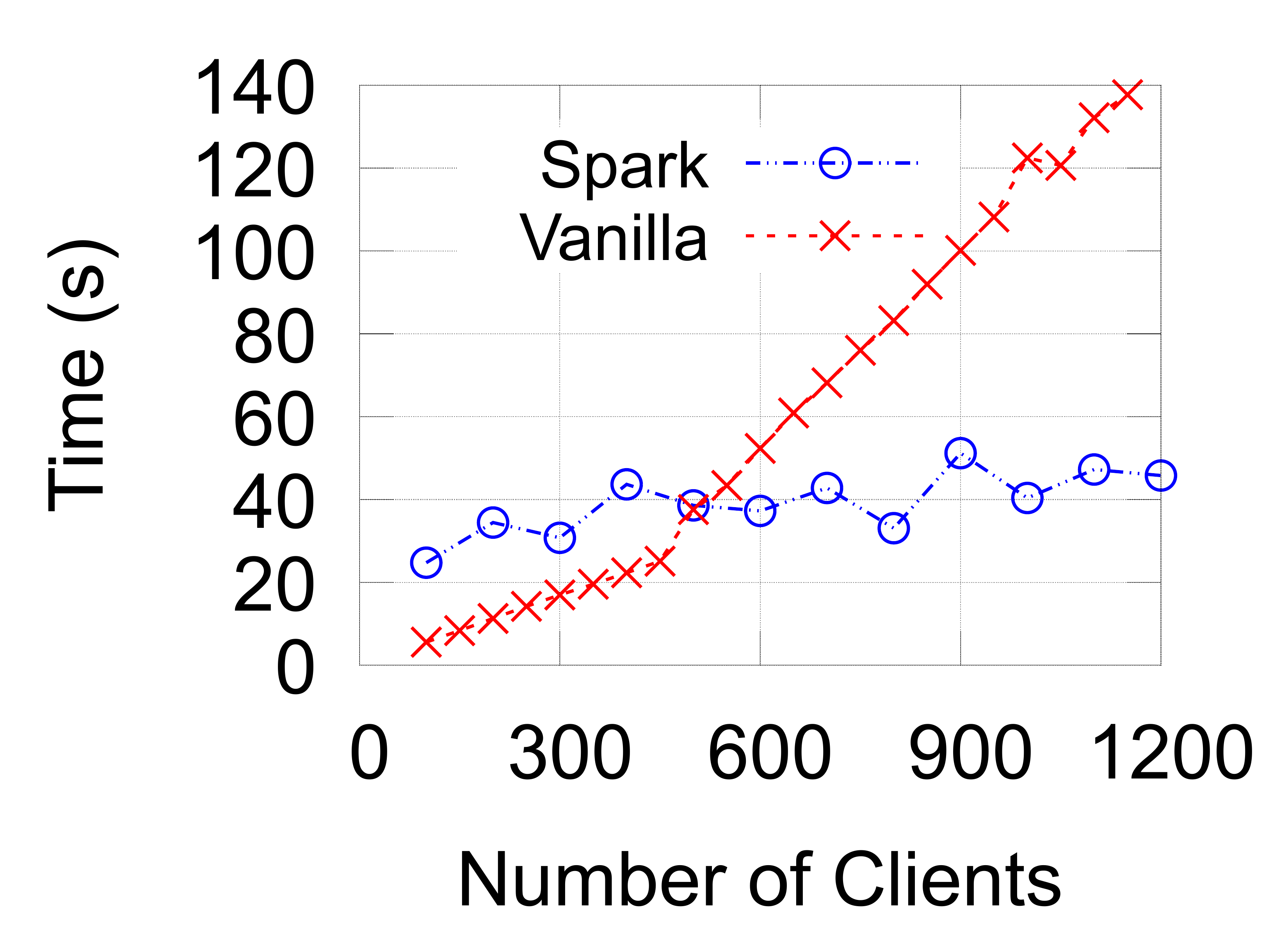}
\caption{CNN73}
\label{fig:pyspark_fedavg_73MB_pyspark_numpy_comparison_no_compressed}
\end{subfigure}
\begin{subfigure}[htbp]{0.21\textwidth}
\centering
\includegraphics[width=1.00\textwidth]{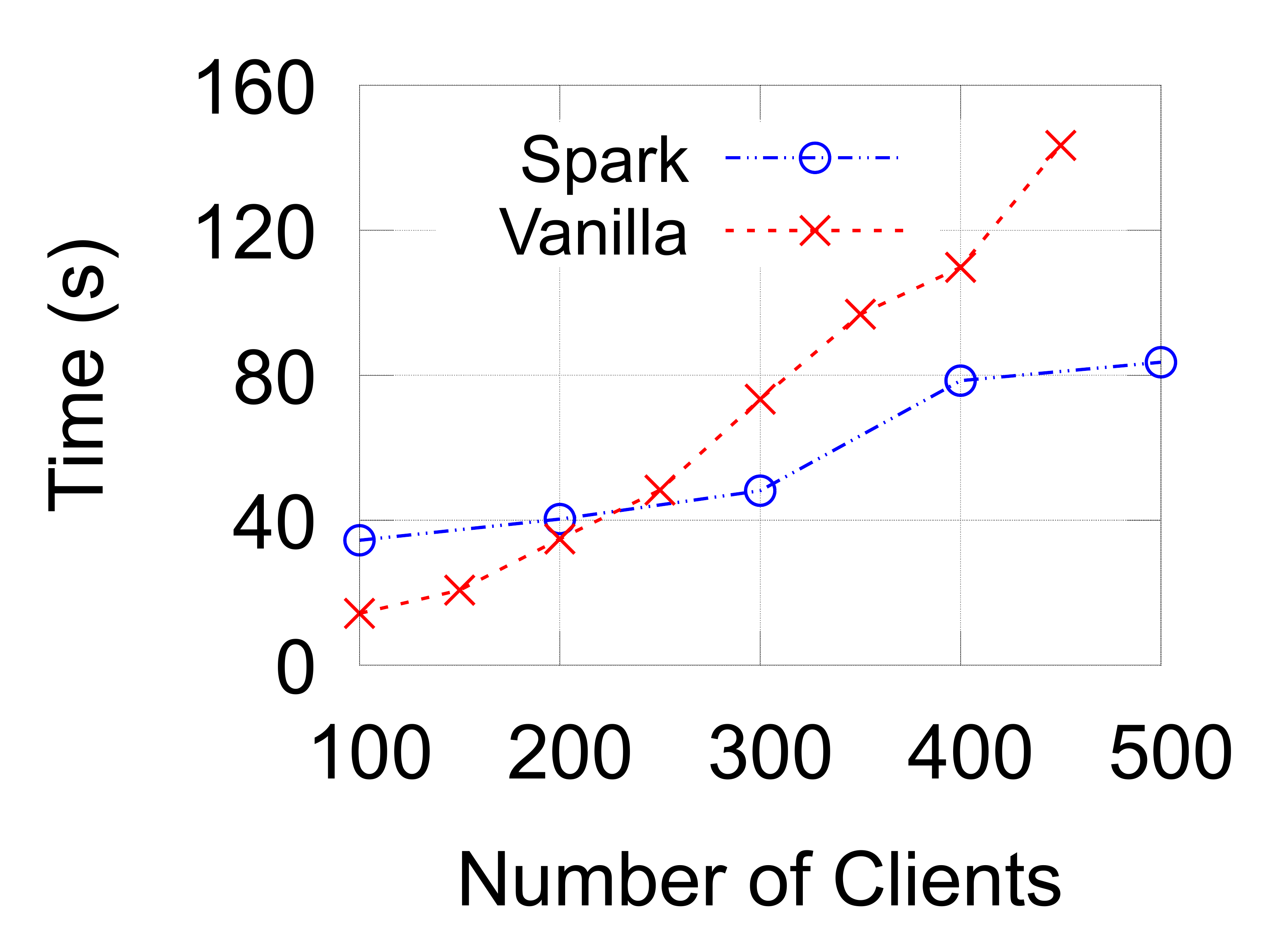}
\caption{CNN179}
\label{fig:pyspark_fedavg_179MB_pyspark_numpy_comparison_no_compressed}
\end{subfigure}
\begin{subfigure}[htbp]{0.21\textwidth}
\centering
\includegraphics[width=1.00\textwidth]{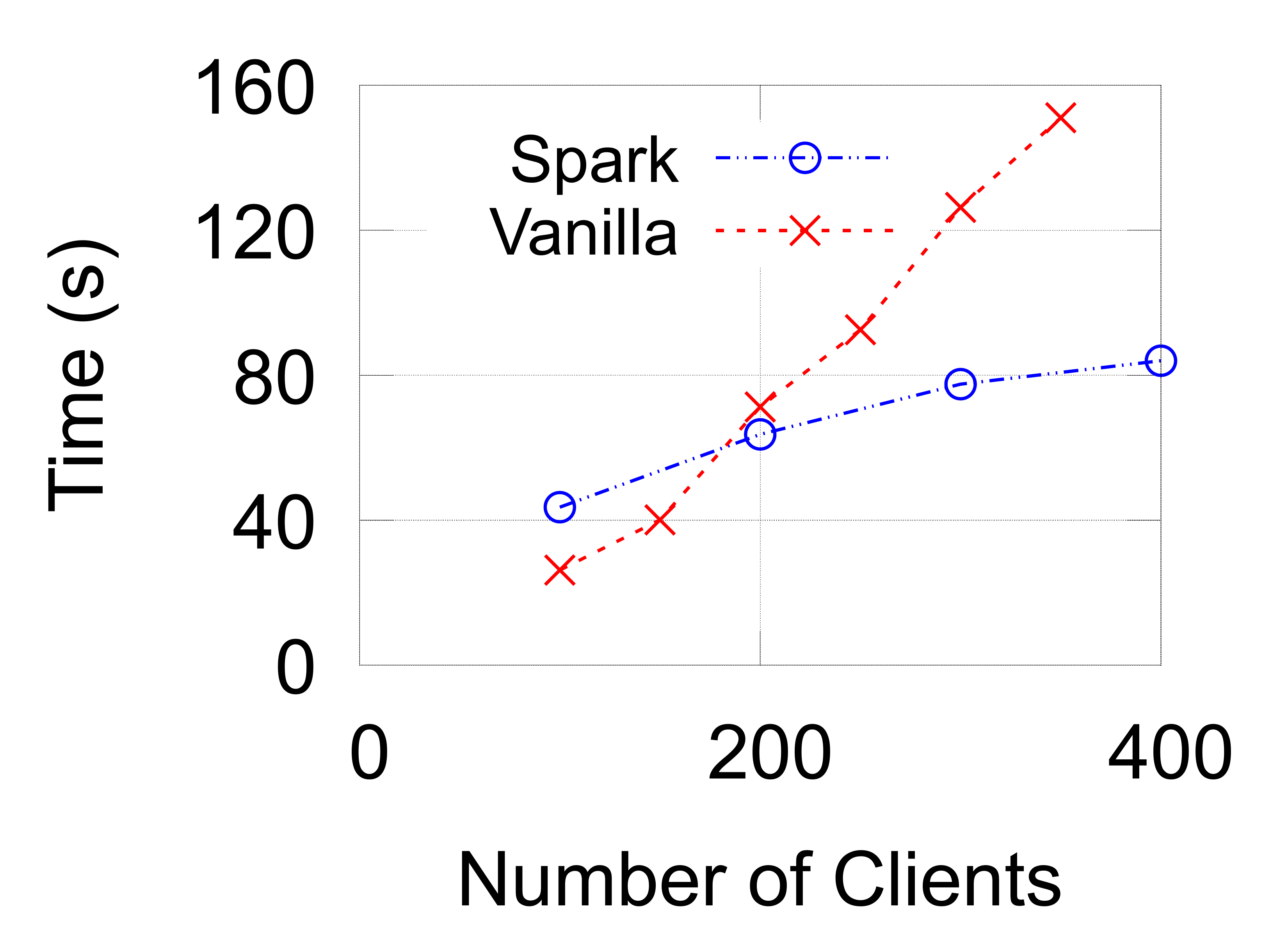}
\caption{CNN239}
\label{fig:pyspark_fedavg_238MB_pyspark_numpy_comparison_no_compressed}
\end{subfigure}
\vspace{-0.5em}
\caption{Average aggregation time comparison for Vanilla and Spark-based method for FedAvg without model compression}
\label{fig:pyspark_numpy_comparison_fedavg_no_compression}
\vspace{-1.5em}
\end{figure*}

\begin{figure}
\centering
\centerline{\includegraphics[width=0.5\columnwidth]{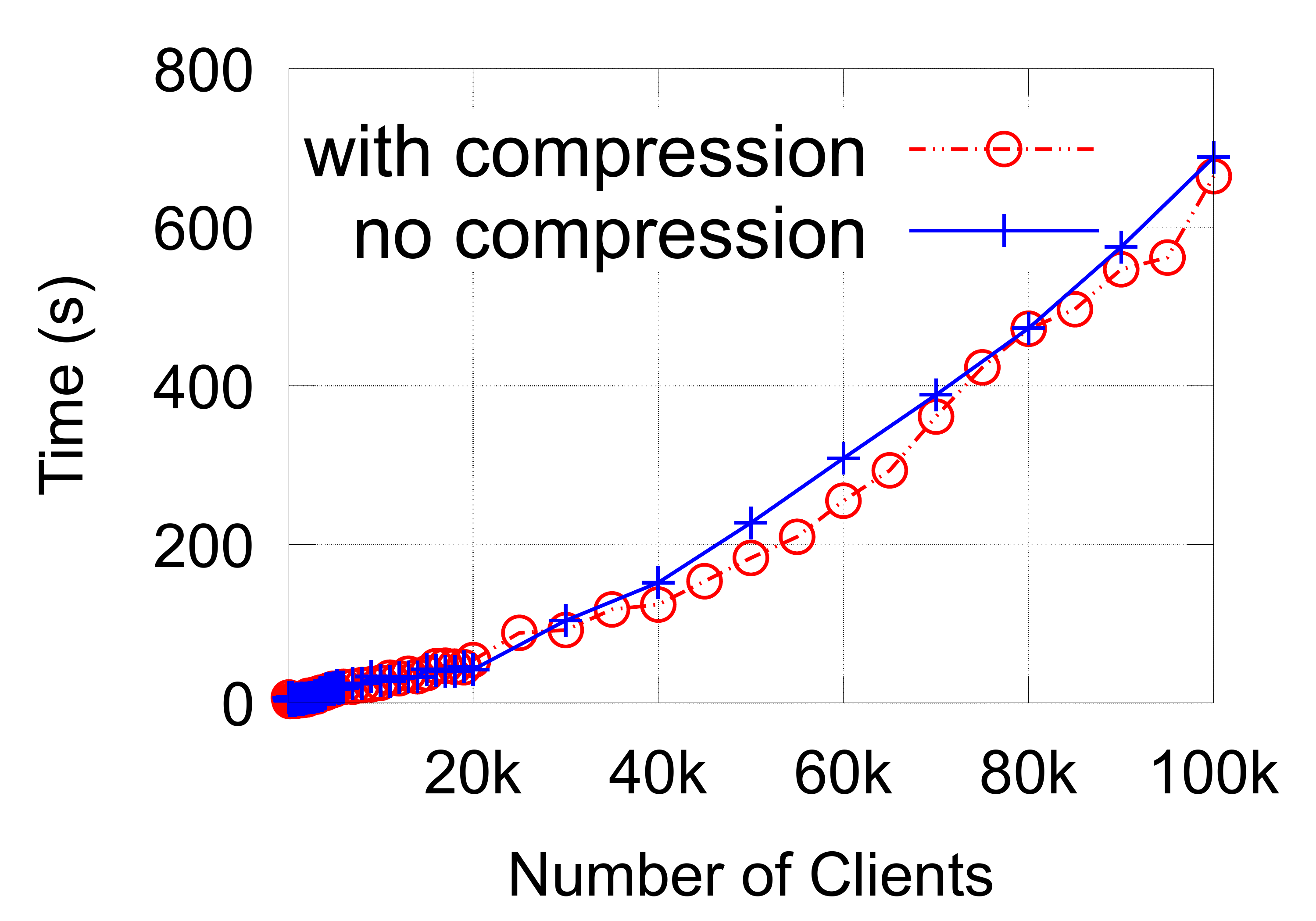}}
\vspace{-8pt}
\caption{Average aggregation time for Spark with FedAvg on CNN4.6 model with and without model compression}
\label{fig:pyspark_fedavg_compression_comparison}
\vspace{-10pt}
\end{figure}

\mysubsection{Spark-based Multi-node Aggregation}
\label{Multi-Node Aggregation}


The Spark-based aggregator in the adaptive aggregation methodology, shown in Figure~\ref{fig:distributed_architecture}, includes a scalable storage monitor. This monitor activates Spark for aggregation after a timeout or a set number of client updates. The threshold is adjustable to mitigate stragglers. This lightweight monitor runs efficiently as a single-threaded process.


The Spark module performs aggregation in the following steps: \ding{202} After each training round, client-sent model updates are stored in Hadoop Distributed File System (HDFS) using the webHDFS Rest API. \ding{203} The monitor waits for a threshold to be reached and triggers the Spark cluster module to initiate aggregation. \ding{204} Spark partitions the data and employs the binary files method to read data as bytes in executor containers. The map function converts RDD bytes into RDDs of the Numpy object type. \ding{205} Lastly, Spark MapReduce processes RDDs, applies the fusion algorithm, and stores updated model weights in HDFS for client access via the WebHDFS API.

In a detailed experiment, we tested the Spark-based method with an increasing number of clients using the FedAvg algorithm and the CNN4.6 model. The results in Figure~\ref{fig:pyspark_fedavg_4.6MB} show "MapReduce Time" as the time for weighted averaging of partitions by this method. We used a "lazy read" approach for efficiency, reading only the partition in use. Smaller models were cached, with most RDDs of model weights cached on worker nodes until the reduction step. However, caching was less effective for larger models due to memory constraints.


\vspace{-0.8em}

\begin{tcolorbox}[left=0mm, right=0mm, top=0mm, bottom=0mm]
IoT and Edge devices have limited network resources~\cite{ IoT_challenges, IoT_breakdown}. Compression techniques can affect computation cost, scalability, and aggregator efficiency. In Figure~\ref{fig:pyspark_numpy_comparison_fedavg}, multi-node approaches prove more effective for aggregating compressed models in IoT and Edge FL, reducing compute cost, network usage, and aggregation time for clients.
\end{tcolorbox}
\vspace{-0.7em}

Figure~\ref{fig:pyspark_federated_and_iteravg_4.6} displays the total time required for the Iteravg method, which involves only two simple steps of sum and division for a mean calculation. The number of clients selected per training round was increased iteratively up to 100k in Figure~\ref{fig:pyspark_federated_and_iteravg_4.6}. Both FedAvg and Iteravg displayed a linear trend in the time required to aggregate, and no limitations were observed in their ability to scale horizontally using this Spark-based method. The number of clients supported per training round increased by $429.1\%$ for FedAvg and $207.7\%$ for Iteravg compared to Vanilla. It is important to note that the figure only shows up to 100k clients, but the adaptive method which includes the Spark-based method has the potential to scale for even more clients and 100k represents only $5\%$ of the clients that are selected from the total available clients per round for aggregation which means the total clients can be as much as 2 million. This analysis provides an answer to the scalability question~\stepnum{2} (raised in section~\ref{sec:background}). Although Spark showed scalability, a detailed analysis of its resource efficiency and costs is necessary to examine its practicality for the edge aggregator, which is performed in section~\ref{sec:AdaptiveAggregationPolicy}.\looseness=-1


\begin{figure}[htbp]
\vspace{-1.0em}
\centering
\begin{subfigure}[htbp]{0.49\columnwidth}
\centering
\includegraphics[width=1.00\columnwidth]{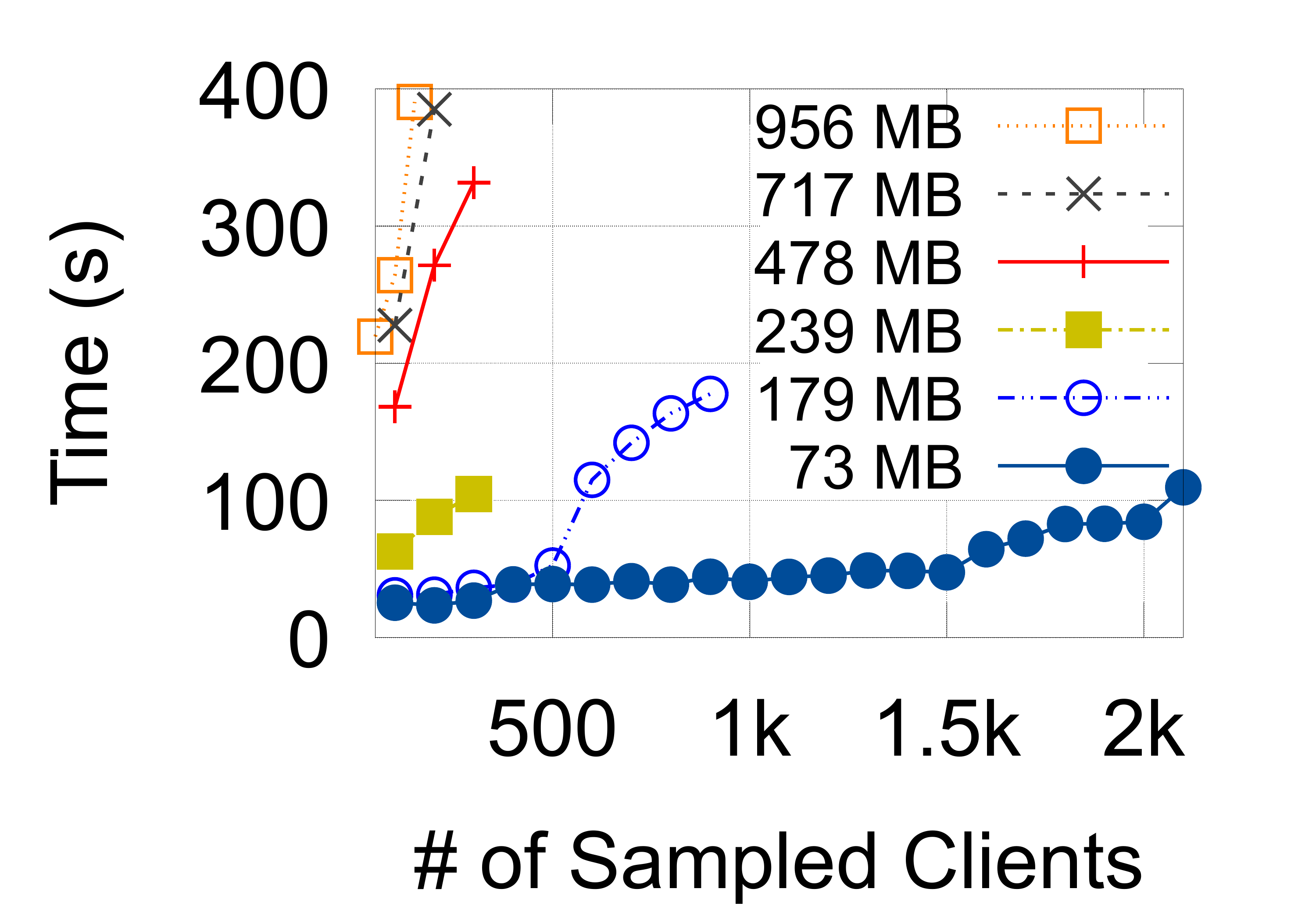}
\caption{Fedavg}
\label{fig:pyspark_fedavg_diff_models}
\end{subfigure}
\begin{subfigure}[htbp]{0.49\columnwidth}
\centering
\includegraphics[width=1.00\columnwidth]{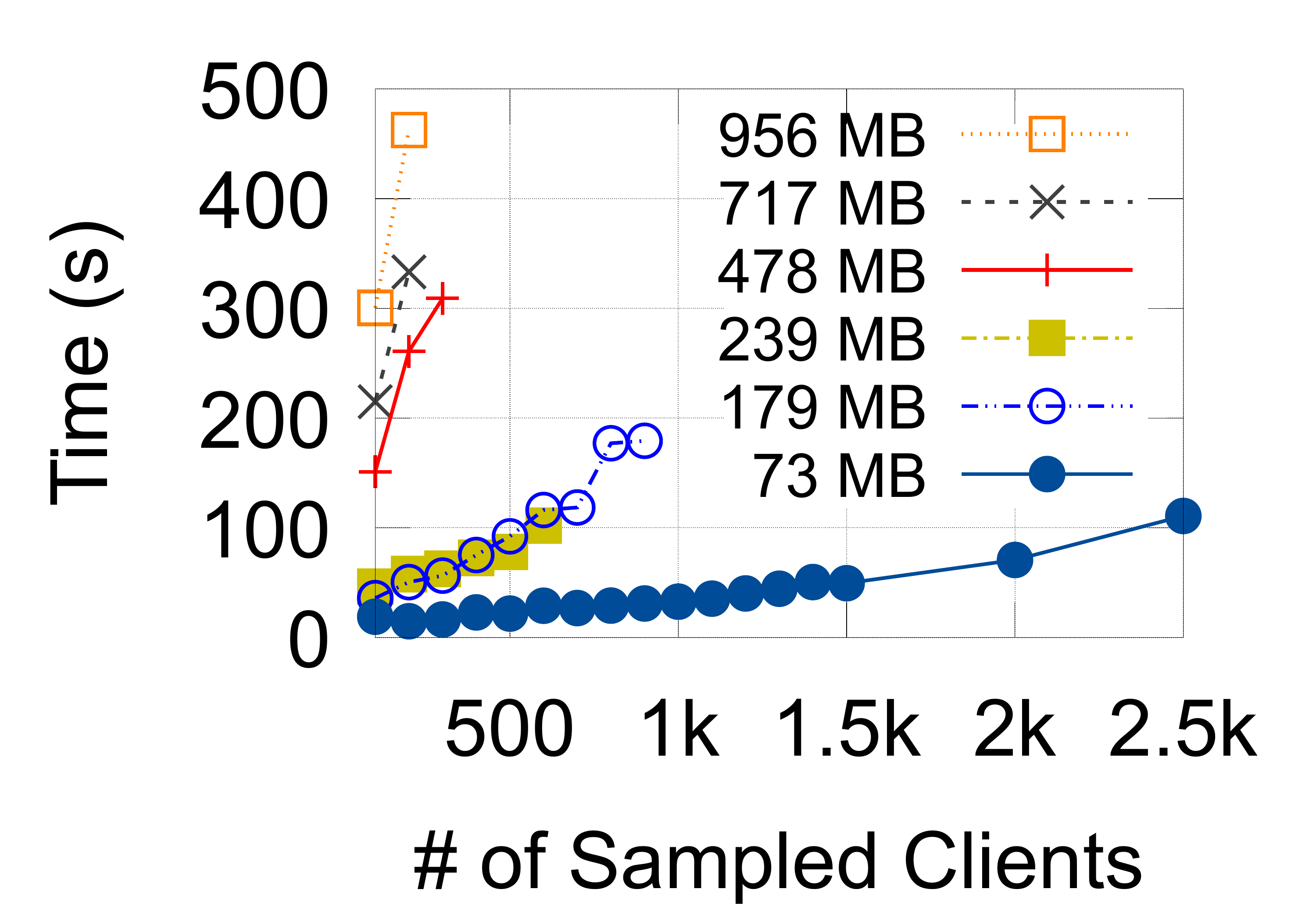}
\caption{Iteravg}
\label{fig:pyspar_iteravg_diff_models}
\end{subfigure}
\vspace{-0.4em}
\caption{Average aggregation time for Fedavg and Iteravg algorithms on varying model sizes with Spark}
\label{fig:pyspark_diff_models}
\vspace{-1.2em}
\end{figure}

We also evaluated various benchmark models that use compressed model updates, which is a technique commonly used to reduce communication costs at edge data centers and IoT devices with limited network bandwidth~\cite{IoT_challenges}. The evaluation results are presented in Figures~\ref{fig:pyspark_numpy_comparison_fedavg} and~\ref{fig:pyspark_numpy_comparison_iteravg_compressed}. The Spark-based method was found to be the most time-efficient when there was a large number of clients, as shown in Figures~\ref{fig:pyspark_fedavg_4.6MB_pyspark_numpy_comparison_compressed} and~\ref{fig:pyspark_iteravg_4.6MB_pyspark_numpy_comparison_compressed}. This is because Spark enables parallel tasks to be executed across multiple executors, thereby increasing the efficiency with higher client participation. To ensure unbiased results, we also conducted the same evaluation without compression. The results, as illustrated in Figure~\ref{fig:pyspark_numpy_comparison_fedavg_no_compression}, exhibit a similar trend. However, the time efficiency gain is more significant compared to Vanilla with compression in Figure~\ref{fig:pyspark_numpy_comparison_fedavg}.

\begin{figure}
\centering
\centerline{\includegraphics[width=1.0\columnwidth]{ 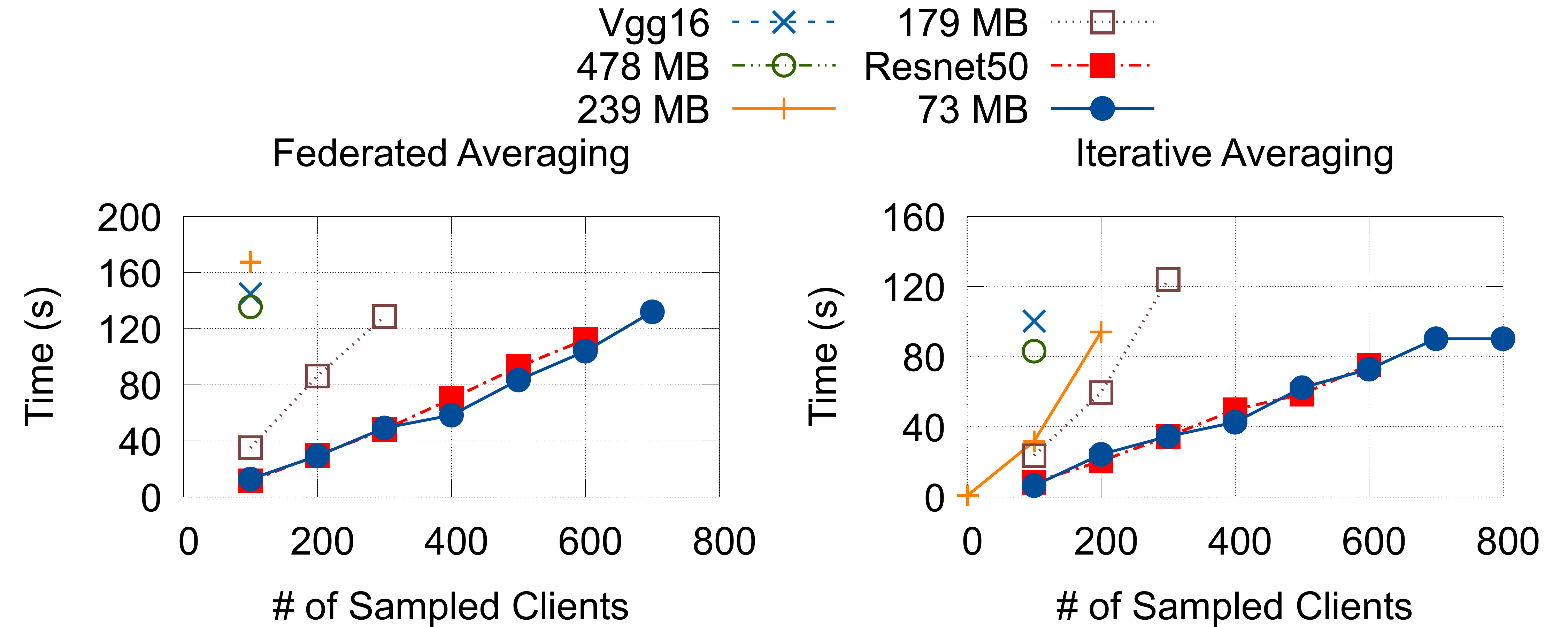}}
\caption{Average aggregation time of Vanilla with varying models \& compression under 170 GB total available memory}
\label{fig:vanilla_aggregation_diff_models_compressed}
\vspace{-16pt}
\end{figure}

In the Spark-based method, the workload is split into multiple workers with separate system resources, reducing the latency involved in the decompression phase as each worker only has to decompress a portion of the total compressed workload. This is more clearly demonstrated in Figure~\ref{fig:pyspark_fedavg_compression_comparison}, where the aggregation latency is similar with or without compressed model updates. In the Vanilla method, decompression with multi-threading is still slow due to the limited memory and CPU resources. This is further evidenced by the Vanilla latency difference between Figures~\ref{fig:pyspark_fedavg_4.6MB_pyspark_numpy_comparison_compressed} and~\ref{fig:pyspark_fedavg_4.6MB_pyspark_numpy_comparison_no_compressed}.\looseness=-1


In summary, while compression increases latency and memory use, thus limiting scalability with Vanilla and Numba methods, it boosts efficiency and scalability with Spark due to its parallel task handling and workload distribution abilities. This makes the Spark-based method more resource-efficient at scale, reducing communication costs at the edge aggregator and maintaining time efficiency.

\vspace{-0.6em}
\begin{tcolorbox}[left=0mm, right=0mm, top=0mm, bottom=0mm]
Figure~\ref{fig:pyspark_numpy_comparison_fedavg_no_compression} shows that Spark works better for heavy models and high client participation surpassing the I/O cost, while Numba is more efficient and cost-effective for lighter models and low participation rates (Sections~\ref{sec:Methodologies} and~\ref{Numba cost analysis}). Thus, an adaptive aggregator is crucial to choose the right method based on IoT and Edge workload and participation rates, ensuring cost-effectiveness.
\end{tcolorbox}

The analysis is extended with compression for different models and increasing numbers of clients in Figure~\ref{fig:pyspark_diff_models}. For Fedavg, the Spark-based method shows a 3X increase in scalability compared to Vanilla. For the CNN73 model, the Spark-based method shown in Figure~\ref{fig:pyspark_fedavg_diff_models} can scale to more than 2.1k clients while Vanilla in Figure~\ref{fig:vanilla_aggregation_diff_models_compressed} only scales to 700 clients under the same evaluation settings described in section~\ref{subsec:Testbed}. Similarly, IterAvg in Figure~\ref{fig:pyspar_iteravg_diff_models} also shows a 3X improvement in scalability compared to Vanilla in Figure~\ref{fig:vanilla_aggregation_diff_models_compressed}.\looseness=-1


\mysubsection{Serverless-based Multi-node Aggregation}
The Serverless method uses the tree-reduce principle and the monitor shown in Figure~\ref{fig:distributed_architecture} to control the multi-level reduction of updates. At each reduction step, the monitor launches multiple Lambda functions concurrently, with the number of updates handled per function dynamically calculated to ensure the memory per function is a maximum of 4 GB. More functions are launched for horizontal scaling to handle additional clients. The Lambda functions take input from the scalable storage used by the Spark module, and the intermediate and final updates are stored in the same scalable storage. In step \ding{204}, the monitor sends AWS Simple Notification Service (SNS) messages to launch Lambda functions for aggregation. Users can adjust the number of Lambda functions and memory limits for each function to optimize efficiency and cost.

\vspace{-0.6em}
\begin{tcolorbox}[left=0mm, right=0mm, top=0mm, bottom=0mm]
Tree-reduce algorithm efficiently distributes workloads in multi-node settings, as demonstrated in Figures ~\ref{fig:pyspark_fedavg_4.6MB_pyspark_numpy_comparison_compressed}, ~\ref{fig:pyspark_iteravg_4.6MB_pyspark_numpy_comparison_compressed}, and ~\ref{fig:pyspark_fedavg_4.6MB_pyspark_numpy_comparison_no_compressed}. This makes multi-core and multi-node aggregator architectures more suitable for Edge and IoT FL applications than Vanilla and other frameworks due to their ability to process workloads in parallel.\looseness=-1
\end{tcolorbox}
\vspace{-0.5em}
\vspace{-0.7em}
\begin{tcolorbox}[left=0mm, right=0mm, top=0mm, bottom=0mm]
For complex workloads in IoT and edge FL applications, Spark excels in efficiency over Serverless, while Serverless is more economical and quicker for simpler models due to reduced per-function costs. Thus, in general, Spark is better for heavier models~\cite{niu2022federated}, and Serverless for lighter ones~\cite{MAROLI2021113488_agriculture}.
\end{tcolorbox}
\vspace{-0.6em}

\begin{figure*}[t]
\begin{subfigure}{1.0\textwidth}
\centering
\includegraphics[width=\textwidth]{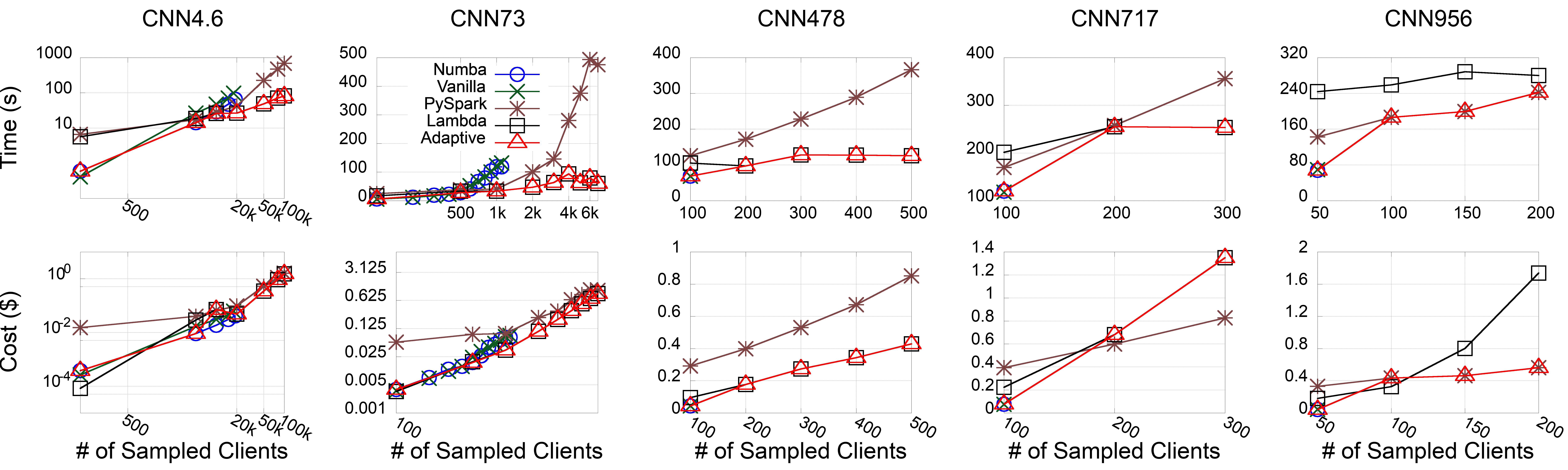}
\phantomcaption
\label{fig:all_techniques_comparison_fedavg_no_compression}
\end{subfigure}

\begin{subfigure}{1.0\textwidth}
\centering
\includegraphics[width=\textwidth]{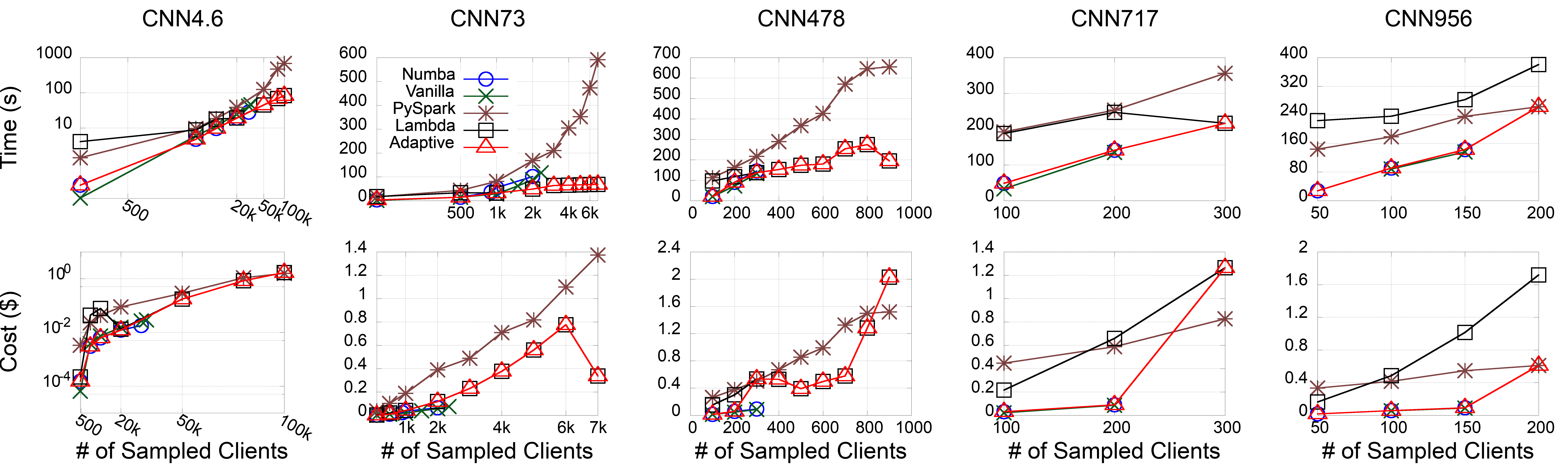}
\phantomcaption
\label{fig:all_techniques_comparison_iteravg_no_compression}
\end{subfigure}
\vspace{-1.5em}
\caption{Comparison of different methodologies for FedAvg (top) and IterAvg (bottom) without compression.}
\vspace{-1.5em}
\label{fig:comparison_no_compression}
\end{figure*}

Experimental analysis shows in Figures~\ref{fig:all_techniques_comparison_fedavg_no_compression} and~\ref{fig:all_techniques_comparison_iteravg_no_compression} that the Serverless method is more efficient and less costly than the Spark-based method for lighter models, but more costly than the Numba-based method when the participation rate of clients is low. Compared to Spark, the Serverless method improves the aggregation efficiency by up to $87\%$ for both FedAvg and IterAvg. This analysis emphasizes the need for an adaptive aggregator, as the Serverless method becomes cost-effective for lighter workloads and can handle unpredictable participation rates of IoT devices, but becomes more costly when the participation rates are lower or the aggregator is dealing with heavier models. The next section provides a more detailed analysis of the Serverless method.

\begin{figure}
\centering
\centerline{\includegraphics[width=0.8\columnwidth]{ 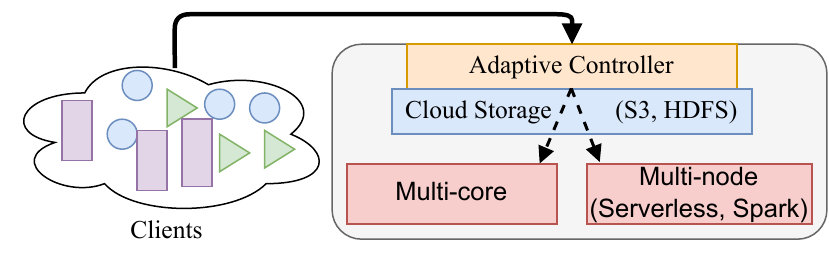}}
\vspace{-10pt}
\caption{Adaptive Aggregator Design}
\label{fig:adaptive_aggregator_design}
\vspace{-15pt}
\end{figure}

\begin{algorithm}[tb]
\caption{RL-based Adaptive Aggregator}
\label{alg:rl_adaptive_aggregator}
    \SetKwInOut{Input}{Input}
    \SetKwInOut{Output}{Output}

        \textbf{Input: }$S_{T}$: Task-specific information, $S_{R}$: System resources information, $\epsilon$: Exploration probability, $\gamma$: Learning rate, $\mu$: Discount factor, $N$: Number of epochs, $T/C$: User preference for time efficiency or cost-effectiveness, if True user prefers efficiency, $P_c$: Penalty factor for completion time, $C_{a}$: Cost of aggregation

    Initialize Q-values $Q(S_{T}, S_{R}, A)$ with random values
    
    \For{$epoch = 1$ \KwTo $N$}{
        Observe state variables $S_{uc}$, $S_{T}$, $S_{R}$
        
        \uIf{random value $< \epsilon$}{
            Select $A$ randomly for exploration
        }
        \Else{
            Choose $A$ with highest $Q$-value for exploitation
        }
        
        Execute aggregation method $A$
        
        Observe completion time $Y_i$
        
        \uIf{$T/C$ is True}{
            $R = -Y_i$
        }
        \Else{
            $R = -C_{a} + P_{c} \cdot Y_i$
        }
        
        Update Q-value: $Q(S_{T}, S_{R}, A) \leftarrow (1 - \gamma) \cdot Q(S_{T}, S_{R}, A) + \gamma \cdot (R + \mu \cdot \max_{A'} Q(S_{T}, S_{R}, A'))$
    }
    
    \textbf{Return} aggregation method $A$ with highest Q-value: $A = \arg\max_A Q(S_{T}, S_{R}, A)$
\end{algorithm}


    
    
    
    

    

    

\section{Adaptive Aggregator}
\label{sec:AdaptiveAggregationPolicy}

This section presents the design of the adaptive aggregator given in Figure ~\ref{fig:adaptive_aggregator_design}, which uses insights from the analysis in section \ref{sec:Methodologies} to develop a predictive model for estimating the completion time of aggregation tasks for each available method. We also conduct an exhaustive analysis of the adaptive method and demonstrate its advantages through a cost, scalability, and efficiency comparison with other methodologies.

We employ a Q-learning-based Reinforcement Learning (RL) agent to optimize the tradeoff between cost, resource utilization, and time efficiency. Let $Y_i$ represent the time taken for a user's aggregation task, which depends on multiple state variables: task-specific information ($S_{T}$), and system resource data ($S_{R}$). Task-specific information includes workload, calculated as the product of the number of clients and the size of model parameters. System information includes available memory, CPU capacity, and the number of executor containers/functions for Spark or Serverless computing. The RL agent learns from these state variables and selects an aggregation method ($A$) to optimize the tradeoff between cost and completion time ($Y_i$), refining its understanding through interactions with the environment. Here's an overview of the algorithm:\looseness=-1

Algorithm~\ref{alg:rl_adaptive_aggregator} optimizes aggregation selection through RL. It initializes Q-values (Line 2), explores or exploits based on Q-values (Lines 3-9), executes aggregation methods (Line 10), observes completion time and calculates rewards based on user preferences (Lines 11-16), and updates Q-values (Line 17). Ultimately, the method with the highest Q-value is chosen for efficiency (Line 19).
Q-value is updated for the chosen action using the Q-learning update equation, incorporating the observed reward, learning rate ($\gamma$), and discount factor ($\mu$). 
The reward calculation takes into account the user's preference for either time efficiency or cost-effectiveness. If $T/C$ is True, it implies the user prefers time efficiency, so the reward is based on negative completion time (-$Y_i$). If T/C is False, indicating the user prefers cost-effectiveness, the reward is calculated by subtracting the cost of aggregation and adding a penalty based on completion time ($P_c$ * $Y_i$). This encourages the agent to minimize completion time while taking into account the user's preference for time efficiency and cost-effectiveness. Hyperparameters $\epsilon$, $\gamma$, and $\mu$ are fine-tuned through sensitivity analysis and experimentation to achieve the desired balances between exploration and exploitation, learning rate, and discount factor.
The adaptive aggregator RL agent optimally adjusts aggregation methods based on Q-values and observed state variables, ensuring efficient tradeoffs between cost, resource use, and time, outperforming other methods in scalability, efficiency, and cost-effectiveness.

\mysubsection{Cost-benefit Study with Adaptive Method}
\label{subsec:ComparativeAnalysis}

The adaptive aggregator can customize cost, scalability, and efficiency requirements for FL aggregators at the edge, making it the first adaptive FL aggregator for Edge and IoT applications. By default, the system chooses the most resource-efficient method based on cost, which is directly translated from resource consumption. This is done while ensuring that the QoS requirements, including time efficiency and scalability, are maintained for the user.

\mysubsubsection{Cost Calculation}
\label{subsubsec:Cost calculation}
The Serverless method's cost is determined by AWS Lambda and depends on function memory and billing duration. Storage costs (S3) are variable and depend on storage characteristics. For the Spark-based method, cost calculation is possible using the pay-to-use AWS Glue API, supporting Serverless ETL operations with PySpark~\cite{sudhakar2018amazon}.

\mysubsubsection{Multi-core in Adaptive Aggregator}
\label{Numba cost analysis}
In fewer-participant IoT applications, both FedAvg and IterAvg start strong with Numba, but face efficiency drops from memory-induced CPU bottlenecks as memory limits are reached, regardless of fusion algorithm complexity, as shown in Figure~\ref{fig:comparison_no_compression} with log scales.




\vspace{-0.7em}
\begin{tcolorbox}[left=0mm, right=0mm, top=0mm, bottom=0mm]
Figure \ref{fig:comparison_no_compression} shows that a fixed multi-node method is overkill for less complex models with low participation rates which makes it more expensive and less efficient than Numba.
\end{tcolorbox}
\vspace{-0.7em}

\begin{table*}
\centering
\caption{Total aggregation time comparison of different aggregation methodologies with the Adaptive method}
\vspace{-0.7em}
\resizebox{0.7\textwidth}{!}{
\begin{tabular}{l|l|l|l|l|}
\cline{2-5}
\multicolumn{1}{c|}{}                                & \multicolumn{1}{c|}{\textbf{Supported Model Size}} & \multicolumn{1}{c|}{\textbf{Scalability}} & \multicolumn{1}{c|}{\textbf{Latency}} & \multicolumn{1}{c|}{\textbf{Average Cost}} \\ \hline
\multicolumn{1}{|l|}{\textbf{Single Node}}           & Very Small                                            & Does not scale                            & Low                                   & \textcolor{green!50!black} {\$}                                          \\ \hline
\multicolumn{1}{|l|}{\textbf{Serverless-TreeReduce}} & Medium                                                & High (except large models)                & Medium                                & \textcolor{green!50!black} {\$\$}                                         \\ \hline
\multicolumn{1}{|l|}{\textbf{Spark}}             & Large                                                 & High                                      & High                                  & \textcolor{green!50!black} {\$\$\$}                                        \\ \hline
\multicolumn{1}{|l|}{\textbf{Adaptive}}                & All                                                   & High                                    & Method dependent                    & \textcolor{green!50!black} {\$\$}                                          \\ \hline
\end{tabular}
}
\label{table:comparison_of_approaches}
\end{table*}

\mysubsubsection{Multi-Node in Adaptive Aggregator}
\label{adaptive cost analysis}

Algorithm \ref{alg:rl_adaptive_aggregator} provides an optimal performance in time and cost, as Figure \ref{fig:comparison_no_compression} illustrates. This adaptive aggregator switches to a Serverless method when client numbers surpass 15k for various CNN models, resulting in significant time and cost savings. For CNN4.6, it reduces latency by 88\% and for CNN73 by 85.59\%, with a \$0.2 cost saving per run compared to a static Spark method. With CNN478, latency decreases by 63.51\%, while for heavier models like CNN956, where Serverless costs increase due to higher cold-start container requirements, the aggregator opts for Spark-based methods, increasing efficiency by 28.73\% and cutting costs by \$1. Although with CNN717 the adaptive method's costs are higher at greater participation levels, favoring efficiency, it can be adjusted to prefer the less expensive Spark-based method, saving \$0.6 each round.

\vspace{-0.6em}
\begin{tcolorbox}[left=0mm, right=0mm, top=0mm, bottom=0mm]
The key point from the analysis in Section ~\ref{sec:Methodologies} is that an adaptive approach is best for addressing diverse workloads in various FL applications. This approach can strike a balance between enhancing efficiency and scalability while also lowering costs through resource-efficient decision-making.
\end{tcolorbox}
\vspace{-0.6em}

Similar patterns are observed with the IterAvg algorithm in Figure ~\ref{fig:comparison_no_compression}, where the adaptive method yields savings in both time and cost. Specifically for workloads like CNN956, the adaptive approach reduces time by 118s and saves over 1 dollar each training round, substantial savings over the hundreds to thousands of rounds in a single FL training job. This algorithm dynamically switches between Numba, Serverless-based, and Spark-based methods, achieving notable cost reductions per round and enhancing QoS, as summarized in Table ~\ref{table:comparison_of_approaches}.

\vspace{-0.6em}
\begin{tcolorbox}[left=0mm, right=0mm, top=0mm, bottom=0mm]
Thundering herd problem can occur if all clients send their data at the same time~\cite{bonawitz2019towards, advances_and_open_problems}, especially with millions of IoT or edge devices with unpredictable participation rates~\cite{mobileNetworks, videoStreamingIoT}, but this effect can be mitigated by using scalable storage (S3/HDFS) with multi-node aggregation.
\end{tcolorbox}
\vspace{-0.3em}

\begin{figure}
\vspace{-1.0em}
\centering
\centerline{\includegraphics[width=0.7\columnwidth]{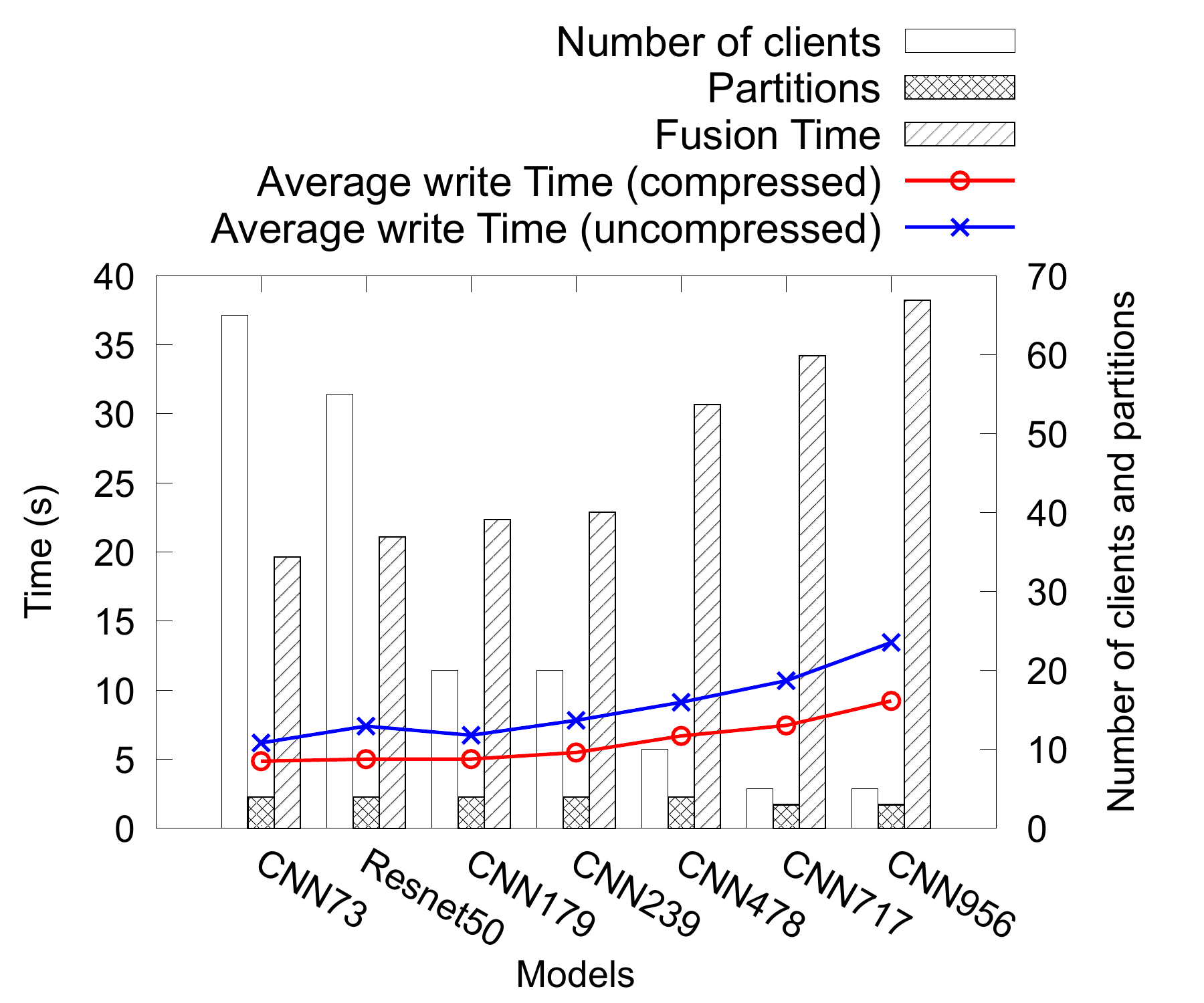}}
\caption{An end-to-end working comparison of the Spark method with simulated clients for varying models using FedAvg (y-axis on the right shows the number of clients and Partitions, y-axis on the left represents all the bars)}
\label{fig:endtoend_diff_models}
\vspace{-1em}
\end{figure}

\emph{Communication in Adaptive Aggregation: }
We assessed the adaptive method's end-to-end latency in small-sized edge data centers~\cite{hierarchical, IoT_challenges}, focusing on the Spark-based method within the adaptive aggregator. Our experiment involved simulated clients on 6 machines connected via a 1 Gigabit Ethernet switch and used scalable HDFS storage for communication (details in section \ref{subsec:Testbed}). To avoid client-side network bottlenecks, we adjusted the number of simulated clients based on network and machine capacities for various model sizes, ensuring the aggregator's write throughput was tested without network constraints on the client side. The results in Figure \ref{fig:endtoend_diff_models} reveal that even with model update sizes increasing by over 9X, the write time only slightly varied. The figure also presents the number of clients, Spark's task partitions for each model aggregation, and the time for reading, writing, and fusion. The average write time refers to the time for writing a single model update from a client, while the reduced time pertains to the MapReduce time for computing the weighted average of partitioned data. We maintained constant partition numbers (Spark tasks) to observe the impact of model size increase on fusion time. Thus, the adaptive aggregator efficiently utilizes scalable storage for client I/O and data input to aggregation task executors (Serverless functions/Spark executors).

\mysubsection{Supplementary Features} 
\label{Discussion}

\begin{table}
\centering
\caption{Aggregation with multiple tenants}
\resizebox{1.0\columnwidth}{!}{%
\begin{tabular}{|l|l|c|l|l|}
\hline
\multicolumn{1}{|c|}{\textbf{Methodologies}} &
  \multicolumn{1}{c|}{\textbf{Models}} &
  \textbf{$\#$ of Sampled Clients} &
  \multicolumn{1}{c|}{\textbf{Cost (\$)}} &
  \multicolumn{1}{c|}{\textbf{Total Time (s)}} \\ \hline
\multirow{2}{*}{Serverless Tree-Reduce} & Resnet50 & 900 & 0.088 & 50.05  \\ \cline{2-5} 
                                        & Vgg16    & 100 & 0.153 & 117.73 \\ \hline
\multirow{2}{*}{Spark MapReduce}        & Resnet50 & 900 & 0.217 & 93.60  \\ \cline{2-5} 
                                        & Vgg16    & 100 & 0.307 & 132.36 \\ \hline
\multirow{2}{*}{Numba}                  & Resnet50 & 400 & 0.019 & 29.94  \\ \cline{2-5} 
                                        & Vgg16    & 80  & 0.038 & 59.74  \\ \hline
\end{tabular}
}
\label{table:multitenant_aggregation}
\vspace{-2.0em}
\end{table}

\begin{table}[h]
\centering
\caption{Emulator write throughput}
\vspace{-0.8em}
\resizebox{1.0\columnwidth}{!}{%
\begin{tabular}{|c|c|c|c||c|c|c|c|}
\hline
\multicolumn{4}{|c||}{\textbf{No drop out}} & \multicolumn{4}{c|}{\textbf{Drop out 5\%}} \\ 
\hline
\textbf{\# Clients} & \textbf{Client Locations} & \textbf{CNN4.6 (s)} & \textbf{CNN478 (s)} & \textbf{\# Clients} & \textbf{Client Locations} & \textbf{CNN4.6 (s)} & \textbf{CNN478 (s)} \\
\hline
1000 & Ireland & 1.99 & 109.77 & 1000 & Ireland & 1.64 & 97.42 \\
& Seoul & 3.401 & 120.84 & & Seoul & 2.55 & 97.58 \\
& California & 1.84 & 98 & & California & 1.66 & 99.65 \\
& Total Time & 13.98 & 238.19 & & Total Time & 3.69 & 116.29 \\
\hline
50000 & Ireland & 1.65 & 114.27 & 50000 & Ireland & 1.8 & 113.83 \\
& Seoul & 5.441 & 126.76 & & Seoul & 5.15 & 125.79 \\
& California & 1.59 & 116.68 & & California & 1.73 & 111.64 \\
& Total Time & 114.48 & 12195.45 & & Total Time & 84.42 & 11670.50 \\
\hline
\end{tabular}
}
\label{table:emulator_eval}
\vspace{-1.5em}
\end{table}

\textbf{Multi-tenant Isolation: } The edge data center's adaptive aggregation service supports multi-tenancy and was tested with VGG16 and Resnet50 models. The results in Table \ref{table:multitenant_aggregation} show that Serverless and Spark performed well with many clients, while the Numba-based method had a limit of 400 clients due to memory constraints. Resource management for multiple tenants is a well-researched topic in cloud services~\cite{2DFQ,workflow_scheduling}, but it's outside our paper's scope.
\textbf{Emulator Evaluation: } 
Table ~\ref{table:emulator_eval} summarizes our emulator evaluation. We tested clients in three global regions, using both small (1000) and large (50k) client numbers, with simpler (CNN4.6) and more complex (CNN478) models. The aggregator server was in Virginia, USA. We measured average client write times and total write times in seconds. Closer locations to the aggregator server (e.g., California) had lower latency, while distant ones (e.g., Seoul) had higher latency due to longer network distances. We also simulated dropouts, randomly dropping 5\% of clients as stragglers, which reduced latency, improving performance. This emulator is valuable for assessing parameter server and FL aggregator performance.
\textbf{Other Multi-node Methods: }
\label{Other distributed methodologies}
We also did an experimental evaluation with Dask, however, it performed less efficiently than Spark due to spending more time on I/O and conversion to its native Bag type. Spark offers better read-and-write throughput with cloud storage and efficiently partitions data for MapReduce computations.
\textbf{Seamless Transition: } 
The adaptive aggregation service seamlessly switches methodologies for different workload sizes to avoid disrupting clients during FL. The I/O channel remains the same in all methods, facilitating smooth transitions. We use the WebHDFS Rest API for HDFS transfers and the Boto3 AWS SDK for Python for S3 storage and retrieval. Clients employ the same APIs for updates. Serverless startup costs are minimal, while Spark context startup costs are hidden during training.
\textbf{Deployment to the Edge: } 
The adaptive aggregator can be deployed at the edge using services like AWS Lambda@Edge~\cite{lambdaedge}, AzureIoTEdge~\cite{azureiotedge}, and OpenEdge~\cite{openedge} for serverless deployment. Numba can run on a simple Linux container, and Spark on Linux-based nodes.
\textbf{Convergence Guarantees: }
In terms of convergence guarantees, the adaptive aggregator ensures the same level of convergence as other systems, as it uses the same fusion algorithm and formula. The difference lies only in the computation technique, without affecting the number of training rounds or final accuracy.
\mysection{Related Work}
\label{RelatedWork}
\emph{Heterogeneity Aware FL: }
Lai et al.~\cite{Oort} proposes Oort, a new participation selection scheme for FL clients. This work looks at improving efficiency from the client's side. It evaluates up to 1.3k clients with small-sized models, out of which 100 are selected by default for aggregation which is approximately 1000X less than the scale we demonstrate. 
\emph{Hierarchical Aggregation: } 
Bonawitz et al.~\cite{bonawitz2019towards} suggest selecting a smaller ratio from available clients to train and creating a hierarchy in the aggregator for scaling but this increases the number of rounds to converge. 
Liu et al.~\cite{hierarchical} suggest a hierarchical FL system in which partial aggregation is done at the edge to distribute load but does not consider the fault tolerance or robustness in the edge and cloud aggregators. 
Having such geographically distributed partial aggregators increases communication time between aggregators and adds extra I/O at each partial aggregator. This work only claims to support up to 1000 lightweight clients in cross-device settings and only a few heavier model clients in cross-silo settings.
\emph{Serverless Aggregation: }
Grafberger et al.~\cite{fedless} propose a serverless solution for both the client and the server. Due to the short-lived nature of Lambda functions it needs to create special provisions to support larger model training and does not mention aggregator scalability.
Jayaram et al.~\cite{lambdaFL} suggest a serverless aggregator which is horizontally scalable with a Kubernetes~\cite{KubernetesManual2017} cluster. This reduces the cost of aggregation by using the pay-to-execute model with serverless functions but has the same limitations as~\cite{fedless}.~\cite{MLSYS2022_f340f1b1} proposes PAPAYA: An asynchronous FL system, however in this paper our main focus is on synchronous FL solutions.
Almost all the fore-mentioned methods assume that there are unlimited network, computation, and memory resources at the Edge, which is unrealistic. In addition, these static techniques are all cloud-based solutions that cannot handle each workload with cost-effectiveness and resource-efficiency, leading to a degradation in the QoS for the user. Furthermore, none of the works tackle all three challenges of scalability, efficiency, and cost reduction.\looseness=-1
\vspace{-0.1em}
\mysection{Conclusion}
\label{Conclusion}
FL is increasingly used in Edge and IoT with edge data center servers to cut communication costs. However, conventional cloud-based aggregators, designed for unlimited resources, face challenges with scalability and efficiency, resulting in higher latency and costs. This study introduces an adaptive aggregator that selects from three methodologies to improve scalability and resource efficiency, and to reduce costs and latency. This adaptive approach also provides users control over cost and efficiency, offering insights into FL aggregation's challenges in edge data centers for IoT and Edge applications, emphasizing the need for flexible solutions.

\bibliographystyle{IEEEtran}
\bibliography{ref}


\end{document}